\documentclass[10pt, a4paper]{article}

\usepackage{lrec2026}

\usepackage{times}
\usepackage{latexsym}
\usepackage[T1]{fontenc}
\usepackage[utf8]{inputenc}
\usepackage{microtype}
\usepackage{inconsolata}
\usepackage{graphicx}
\usepackage{tabularx}
\usepackage{amsmath}
\usepackage{amssymb}
\usepackage{cleveref}
\usepackage[dvipsnames,table]{xcolor}
\usepackage{multirow}
\usepackage{tabularx, colortbl}
\usepackage{paralist}
\usepackage{booktabs}
\usepackage{subcaption}

\usepackage{tikz}
\usetikzlibrary{backgrounds, positioning, fit, calc}
\usepackage{pgfplots}
\pgfplotsset{compat=1.17}

\usepackage{diagbox}

\newcommand\compacteq{\mkern1.5mu{=}\mkern1.5mu}

\colorlet{outerblockscolor}{Cerulean!25}
\colorlet{innerblockscolor}{Cerulean!10}
\colorlet{bmcolor}{Orange!25}
\colorlet{mmcolor}{Orchid!25}
\colorlet{mm2color}{GreenYellow!25}

\colorlet{baselinecolor}{Blue!90}
\colorlet{pcrcolor}{BurntOrange}

\colorlet{angercolor}{BrickRed!75}
\colorlet{boredomcolor}{gray!90}
\colorlet{disgustcolor}{Green!80}
\colorlet{fearcolor}{purple!80}
\colorlet{guiltcolor}{brown!80}
\colorlet{joycolor}{Dandelion!80}
\colorlet{pridecolor}{Apricot!80}
\colorlet{reliefcolor}{cyan!80}
\colorlet{sadnesscolor}{blue!80}
\colorlet{shamecolor}{Lavender!80}
\colorlet{surprisecolor}{Orange!80}
\colorlet{trustcolor}{Blue!80}
\colorlet{noemotioncolor}{black!80}

\makeatletter
\renewcommand\paragraph{\@startsection{paragraph}{4}{\z@}%
  {0.8ex \@plus1ex \@minus.2ex}%
  {-1em}%
  {\normalfont\normalsize\bfseries}}
\makeatother

\title{Disambiguation of Emotion Annotations by \\ Contextualizing Events in Plausible Narratives}

\name{Johannes Sch\"afer, Roman Klinger}

\address{Fundamentals of Natural Language Processing, University of Bamberg, Germany\\
  \texttt{\{johannes.schaefer,roman.klinger\}@uni-bamberg.de}\\}

\abstract{
Ambiguity in emotion analysis stems both from potentially missing information and the subjectivity of interpreting a text.
The latter did receive substantial attention, but can we fill missing information to resolve ambiguity?
We address this question by developing a method to automatically generate reasonable contexts for an otherwise ambiguous classification instance.
These generated contexts may act as illustrations of potential interpretations by different readers, as they can fill missing information with their individual world knowledge.
This task to generate plausible narratives is a challenging one: We combine techniques from short story generation to achieve coherent narratives.
The resulting English dataset of \textsc{E}motional \textsc{B}ack\textsc{S}tories, \textsc{EBS}, allows for the first comprehensive and systematic examination of contextualized emotion analysis.
We conduct automatic and human annotation and find that the generated contextual narratives do indeed clarify the interpretation of specific emotions.
Particularly relief and sadness benefit from our approach, while joy does not require the additional context we provide.
\\ \newline \\ }

\begin{document}

\maketitleabstract

\section{Introduction}\label{sec:intro}
\begin{table}[tbp]
	\centering
	\small
	\fontsize{9.5pt}{9.5pt}\selectfont
	\renewcommand{\arraystretch}{1.}
	\begin{tabularx}{\linewidth}{@{\hspace{1.75mm}}X@{\hspace{1.75mm}}}
		\toprule
		\textbf{Guilt:} A close friend entrusted me with setting up the sound system for a charity event. I faced a daunting task, as the previous technician had left the equipment in disarray. With time running out, I took a shortcut to meet the deadline, skipping some crucial safety checks. Just before the event started, my friend reminded me of its significance and the many people counting on its success. \textit{The loudspeaker suddenly malfunctioned and went silent.}\\
		\textbf{Relief:} I was tasked with giving a presentation to a large crowd. The sound system malfunctioned, amplifying my voice to an ear-piercing level. The sound technician ignored the problem and chatted with someone. The audience covered their ears and looked at me with discomfort. \textit{The loudspeaker suddenly malfunctioned and went silent.}\\
		\textbf{Fear:} I arrived at a remote wilderness survival training camp, where the instructors emphasized the importance of following loudspeaker instructions for safety. The instructors warned us about the toxic waste site nearby and explained that the loudspeaker would alert us to any changes in air quality. During the first exercise, I struggled to navigate the challenging terrain, but the loudspeaker provided crucial guidance, helping me stay on track. I completed a difficult obstacle course, relying heavily on the loudspeaker's instructions to avoid hazards and find the safest route. \textit{The loudspeaker suddenly malfunctioned and went silent.}\\
		\textbf{Pride:} I spent the entire morning upgrading the sound system with a new backup system to prevent technical issues. The event organizer informed me that the conference was running 30 minutes behind schedule, giving me extra time to test the new backup system. I used the extra time to run a series of tests on the sound system, trying to simulate potential failures. The keynote speaker began to talk, and the sound system was working flawlessly, but I was still waiting for a real test of the new backup system. \textit{The loudspeaker suddenly malfunctioned and went silent.}\\
		\bottomrule
	\end{tabularx}
	\caption{Example narratives for different emotions.}
	\label{tab:examples}
\end{table}

When interpreting event descriptions, ambiguity in emotion analysis can arise for various reasons.
Studies have attempted to explain low inter-annotator agreement by differences in demographics among annotators, which can contribute to varied interpretations in subjective tasks  \citep[i.a.]{Wan_Kim_Kang_2023,mieleszczenko2023capturing,plaza-del-arco-etal-2024-emotion,sun-etal-2025-sociodemographic}.
Factors such as age, cultural background, and personal experiences may influence the interpretation of emotion responses to an event.

However, what has received no attention so far is the impact of variations in the textual context of event descriptions, which can also significantly shape emotion interpretations and responses.
For instance, given the following text from the dataset\footnote{We provide our annotated dataset and code on \url{https://www.uni-bamberg.de/en/nlproc/resources/emotional-backstories/}.} we present in this paper: \textit{``The loudspeaker suddenly malfunctioned and went silent.''}, a prevalent interpretation may be that this event causes surprise.
However, the context of the event may change the distribution of possibly experienced emotions as shown in \Cref{tab:examples}.
In the case of guilt, the added narrative establishes a profound sense of personal responsibility and the anticipated negative consequences of actions, which are then triggered by the event.
Relief emerges from including a backstory that portrays a situation initially fraught with anxiety and possibly negative consequence beyond one's control, suggesting that these are resolved by the event.
Fear is framed by a narrative that expresses the unpredictability of a challenging circumstance, emphasizing a sense of uncertainty and the immense effort required to navigate the unknown, reinforcing an unpleasant loss of control based on the event.
Lastly, pride is evoked by incorporating a backstory that highlights proactive efforts and successful task management, illustrating how these accomplishments will be recognized by others as a result of the event.
These examples demonstrate that backstories elicit cognitive appraisals for each response, which lead to emotion interpretation differences in the event.
We assume that readers of ambiguous events similarly fill this context based on their world knowledge.

While there have been studies on emotion classification under the paradigm of perspectivism \citep[i.a.]{milkowski-etal-2021-personal,suzuki-etal-2022-emotional,KAZIENKO202343,troiano2023dimensional}, we are not aware of any work that studies the importance of context to disambiguate emotion analysis in a controlled manner.\footnote{Emotion recognition in conversations does consider context, but not to disambiguate otherwise unclear utterances in a controlled manner \citep[i.a.]{hu-etal-2021-dialoguecrn}.}
Ambiguity in the interpretation of event descriptions viewed in isolation can stem from missing essential information, which can also lead to uncertainty in emotion responses.
Furthermore, not all event descriptions elicit distinct emotions; some are devoid of specific emotion, and thus can be filled only through narratives.
We approach disambiguation by generating different contexts as sequences of preceding events, thereby examining how various backstories can have different effects on the emotion interpretation of that event.
Our proposed systems generate event chains by prompting a large language model (LLM) in different settings, with the goal to understand whether these contexts lead to a higher agreement in human annotation of emotions.

In our experiments, we first assess the quality of our generated data, to ensure that it is suitable for an evaluation of human annotation and automatic predictions, which is our main goal.
Hence, our paper is guided by the following research questions:
\newcommand{\RQone}{Does context generation through iterative prompting with story planning enhance narrative coherence or is a one-step prompting approach sufficient? }
\newcommand{\RQtwo}{Do generated contexts enhance clarity in human annotation of emotions in events? }
\newcommand{\RQthree}{Can systems recognize the contextual disambiguation in emotion analysis of events? }
\begin{compactenum}
	\item \RQone
	\item \RQtwo
	\item \RQthree
\end{compactenum}

To answer the questions, we construct a dataset (\textsc{E}motional \textsc{B}ack\textsc{S}tories, \textsc{EBS}) as the foundation to study the influence of contextual information on emotion prediction.
Our work is situated at the intersection of explainable artificial intelligence (XAI) and perspectivism. Our contribution fits to XAI, because the generated context functions as an explanation of a possible textual interpretation. It is related to perspectivism, because multiple such contexts are reasonable.

In the remainder of this paper, we review related work in \Cref{sec:rw}, present our data generation and analysis methods in \Cref{sec:datagen}, and analyze our experimental results in \Cref{sec:data}.

\section{Related Work}\label{sec:rw}
The novel task of generating backstories that evoke specific emotions in events intersects with multiple research fields, which we review in this section.

\subsection{Contextual Influences on Event Interpretations}
The interpretation of text is significantly shaped by its context.
\citet{das2011temporal} explore the dynamics of emotions in their analysis of event chains, with the goal of finding the correct temporal sequencing of events.
They focus on identifying the sentiments and emotions associated with events and utilize these as features to uncover contextual relationships.
\citet{mostafazadeh-etal-2016-corpus} introduce the ROCstory dataset, which comprises crowd-sourced narratives, each consisting of a series of five events.
This corpus is intended to assist in story completion tasks, where a system must determine the most fitting event to follow the first four provided events.
\citet{9070506} conduct sentence-level emotion analysis on this dataset, although they do not take prior contextual influences into account.
The GLUCOSE dataset introduced by \citet{mostafazadeh-etal-2020-glucose} further enhances understanding of complex dependencies between events through commonsense inferences; however, it does not specifically address the emotions being evoked in these contexts.

\subsection{Emotion Categorization in Context}
Emotion analysis seeks to identify emotion states elicited in readers or authors of texts.
The foundational frameworks by \citet{ekman1972universals} and \citet{Plutchik2001} provide categorization methodologies.
\citet{mohammad-2012-emotional} created emotion-labeled social media corpora, while \citet{troiano2023dimensional} emphasized implicit emotion cues linked to appraisal theories.
Their analysis of event descriptions without context highlights a gap our study addresses by focusing on contextual influences.
Notably, \citet{etienne2022emotionannotation} emphasize the intricacies of emotion expression across diverse texts, and \citet{etienne2024emotionprediction} predict various modes of emotion expression, emphasizing the complexity that this context brings to emotion annotation.
\citet{chatterjee-etal-2019-semeval} and \citet{wemmer-etal-2024-emoprogress} advocate for considering prior context in dialog systems to refine emotion interpretations.
The dynamic interplay between event components and emotions is also acknowledged by \citet{cortal2023emotionalnarratives}.
Additionally, \citet{Labat2024} investigate emotion trajectories in customer service dialogues, emphasizing that emotions may shift with each utterance.
Our work enhances these studies with an in-depth analysis of contextual influences on emotion analysis in a controlled dataset tailored for this purpose.

\subsection{Story Generation Techniques}
Story generation methodologies central to our study involve techniques designed to enhance narrative quality and coherence.
\citet{razumovskaia-etal-2024-little} emphasize the benefits of story planning, while \citet{ma-etal-2023-coherent} examine how integrating structured knowledge can improve coherence.
\citet{xie-riedl-2024-creating} and \citet{NEURIPS2023_ae9500c4} demonstrate effective planning and diversification strategies, and \citet{hosseini-etal-2024-synthetic} focus on generating high-variance datasets for natural language inference tasks.
Additionally, \citet{chung-etal-2023-increasing} analyze the balance between accuracy and diversity of generated data.
\citet{yang2024makesgoodstorymeasure} address challenges in establishing effective criteria for evaluating story quality, while \citet{long-etal-2024-llms} discuss the need for human intervention to maintain data faithfulness and diversity.
Furthermore, \citet{eigenschink2023deep} and \citet{li-etal-2023-synthetic} emphasize the importance of generating synthetic data that adheres to realism, diversity, and coherence. 
These varied insights guide the generation approach in our setting with regard to narrative quality and relevance.

\subsection{Backstory Generation}
\citet{clark2022working} explore character backstories in game design with a multi-step approach to develop engaging narrative complexities.
\citet{moon2024virtualpersonaslanguagemodels} produce realistic persona backstories using open-ended prompts, while \citet{stricker-paroubek-2024-chitchat} suggest including specific instructions in prompts in order to create user backstories for enhanced conversational realism.
Furthermore, \citet{jung-etal-2023-enhancing} focus on generating preceding dialogue turns to augment user-system interactions with a focus on cohesive alignment.

Overall, the literature emphasizes the necessity of coherent, contextually diverse narratives.
Existing research tends to generate free-form stories or focus on story completion tasks, where the emphasis is on crafting endings for pre-existing narratives.
In contrast, our study aims to generate backstories that influence specific events, necessitating a novel reverse construction method.
Insights into prompt setups, stepwise approaches, content diversification, story planning, and evaluation criteria for narrative quality further inform our methodology.

\section{Generation of Event Chains} \label{sec:datagen}
In order to systematically perform contextualized emotion analysis, we generate short stories that contain narratives according to specific requirements with an LLM, considering various settings.
We now define the underlying concepts, outline the procedural steps of our generation approaches, and present our evaluation methods.

\subsection{Definitions}\label{sec:def}
\paragraph{Emotion Modeling.}
In our study, emotion analysis of text refers to modeling a person's interpretation of the events they describe.
Following  \citet{troiano2023dimensional}, we use a fine-grained emotion model consisting of $n\compacteq13$ categories: $\mathbb{E} \compacteq \{e_1, \ldots, e_n\} \compacteq \{$anger, boredom, disgust, fear, guilt, joy, pride, relief, sadness, shame, surprise, trust, no-emotion$\}$.
The rationale for selecting these categories stems from their grounding in appraisal processes, which provide vital insights into emotion responses.
We suspect that relevant appraisal processes may be articulated within the narrative contexts we aim to generate.
Moreover, by employing a larger set of emotions, we allow for a fine-grained analysis that can lead to richer and more diverse texts.
Broader analyses can be performed by grouping of some finer categories into overarching classifications, such as distinguishing between positive and negative emotions.

\paragraph{Event Chains.}
We generate short stories as event chains consisting of five distinct sentences $c \compacteq \{s_1, ..., s_5\}$, which is in line with data in the corpus ROCstory \citep{mostafazadeh-etal-2016-corpus}.
Each sentence contains a description of a singular event from the perspective of a person experiencing it.
The five events in a chain are organized chronologically to form a coherent narrative.
We consider the first four events in a chain as the backstory $b \compacteq \{s_1, \ldots, s_4\}$ and analyze its influence on the emotion being evoked in the event experiencer in the concluding event $e(b \cdot s_5)$.
To facilitate a comprehensive analysis of diverse emotion responses, we associate multiple distinct backstories with the same concluding event $s_5$.
For our dataset, we aim to include precisely one backstory for each emotion category for each concluding event: $\mathbb{B}(s_5) = \{ b_i \mid e(b_i \cdot s_5) \in \mathbb{E}\}$ with $i \compacteq 1, \ldots, n$.

\subsection{Generation Approaches}\label{sec:gens}
\begin{figure}
	\centering
	\begin{tikzpicture}[]
		
		\node (block1) [rectangle, minimum width=.95\linewidth, text width=.925\linewidth, inner sep=0pt] {\small\textsf{I. Generation of Events}};
		\node (innerblock0) [rectangle, fill=innerblockscolor, minimum width=.45\textwidth, text width=.45\textwidth, below of=block1, node distance=.4cm, inner sep=2pt] {\scriptsize \textsf{Prompt 1: Describe event given type and object.}};
		
		\begin{pgfonlayer}{background}
			\node[fit=(block1)(innerblock0), draw=none, fill=outerblockscolor, inner sep=2pt,] {};
		\end{pgfonlayer}
		
		\tikzstyle{methodbox} = [rectangle, draw, thick, below of=innerblock0, node distance=2.35cm, rounded corners=5pt, minimum width=2.35cm, minimum height=3.075cm]
		\node (bm) [methodbox, xshift=-2.5cm, fill=bmcolor] {};
		\node (mm) [methodbox, fill=mmcolor ] {};
		\node (mm2) [methodbox, xshift=2.5cm, , fill=mm2color] {};
		
		\node (block2) [draw=none, minimum width=.95\linewidth, text width=.925\linewidth, node distance=.8cm,below of=innerblock0, inner sep=0pt] {\small\textsf{II. Generation of Backstories as Event Chains}\newline \textsf{Baseline:} \hspace*{.9cm} \textsf{PC:}\hspace*{1.85cm} \textsf{PCR:}};
		
		\node (innerblock1b) [rectangle, fill=innerblockscolor, minimum width=.27\linewidth, minimum height=.5cm, text width=.26\linewidth, below of=block2, xshift=-2.5cm,  node distance=1.1cm, inner sep=2pt] {\scriptsize\shortstack[l]{\textsf{Prompt 2:}\\ \textsf{Construct chain}\\ \textsf{for emotion and}\\ \textsf{event.}}};
		
		\node (innerblock1) [rectangle, fill=innerblockscolor, minimum width=.27\linewidth, minimum height=.5cm, text width=.26\linewidth, below of=block2, xshift=0.0cm, node distance=.85cm, inner sep=2pt, align=left] {\scriptsize \shortstack[l]{\textsf{Prompt 2.1:}\\ \textsf{Plan story.}}};
		\node (innerblock2) [rectangle, fill=innerblockscolor, minimum width=.27\linewidth, minimum height=.5cm, text width=.26\linewidth, below of=innerblock1, node distance=.95cm, inner sep=2pt, align=left] {\scriptsize \shortstack[l]{\textsf{Prompt 2.2:}\\ \textsf{Construct chain}\\ \textsf{from story.}}};
		
		\node (innerblock12) [rectangle, fill=innerblockscolor, minimum width=.27\linewidth, minimum height=.5cm, text width=.26\linewidth, below of=block2, xshift=2.5cm, node distance=.85cm, inner sep=2pt, align=left] {\scriptsize \shortstack[l]{\textsf{Prompt 2.1:}\\ \textsf{Plan story.}}};
		\node (innerblock22) [rectangle, fill=innerblockscolor, minimum width=.27\linewidth, minimum height=.5cm, text width=.26\linewidth, below of=innerblock1, xshift=2.5cm, node distance=.95cm, inner sep=2pt, align=left] {\scriptsize \shortstack[l]{\textsf{Prompt 2.2:}\\ \textsf{Construct chain}\\ \textsf{from story.}}};
		\node (innerblock3) [rectangle, fill=innerblockscolor, minimum width=.27\linewidth, minimum height=.5cm, text width=.26\linewidth, below of=innerblock22, node distance=.925cm, inner sep=2pt] {\scriptsize \shortstack[l]{\textsf{Prompt 2.3:}\\ \textsf{Revise chain.}}};
		
		\begin{pgfonlayer}{background}
			\node[fit=(block2)(innerblock1)(innerblock2)(innerblock3), draw=none, fill=outerblockscolor, inner sep=2pt] {};
		\end{pgfonlayer}
		
	\end{tikzpicture}
	
	\caption{Overview of our LLM-based data generation framework for event chains according to three different methods. The prompts used in each case are shown summarized -- for full text prompts see \Cref{app:prompts}.}
	\label{fig:overview}
\end{figure}
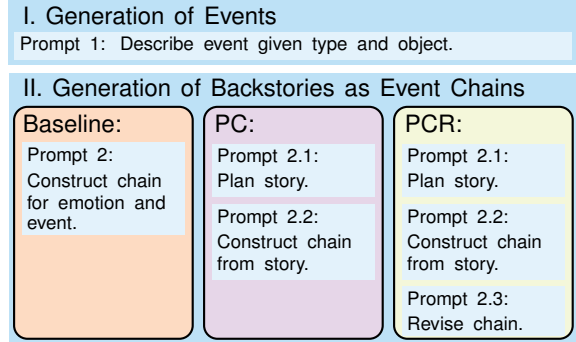

\begin{table*}[t!]\centering
	\centering\sffamily\small
	\renewcommand{\arraystretch}{1.2}
	\begin{tabularx}{\linewidth}{p{5mm}p{3cm}p{2.75cm}X}
		\toprule
		Step&\multicolumn{1}{c}{Input}&\multicolumn{1}{c}{Summarized Prompt}&\multicolumn{1}{c}{Output}\\
		\cmidrule(r){1-1}\cmidrule(rl){2-2}\cmidrule(rl){3-3}\cmidrule(l){4-4}
		I.
		& 
		Event type: \textit{Competition},\newline
		Event type object: \textit{communication tools}.
		& 
		Prompt 1: Describe event given type and object.
		& 
		\textit{The loudspeaker suddenly malfunctioned and went silent.}\\
		\cmidrule(r){1-1}\cmidrule(rl){2-2}\cmidrule(rl){3-3}\cmidrule(l){4-4}
		II.
		& 
		Event: \textit{The loudspeaker suddenly malfunctioned and went silent.},\newline
		Emotion: \textit{Guilt}.
		& 
		Prompt 2: Construct backstory for emotion and event.
		& 
		I had been tasked with testing the loudspeaker system before the big event.
		My supervisor warned me that a malfunction would be disastrous for the company's reputation.
		I skipped the recommended final check to grab a quick lunch before the event started.
		The event host began the ceremony, and the loudspeaker was working perfectly, filling me with temporary relief.
		\textit{The loudspeaker suddenly malfunctioned and went silent.}\\
		\bottomrule
	\end{tabularx}
	\caption{Steps for target event generation and backstory generation (Baseline method) on an example.}\label{tab:examplesg}
\end{table*}

We generate diverse event descriptions (the set of events $s_5$) and then compare three methods for generating contexts for each of them, as visualized in \Cref{fig:overview}.
The generation process, based on the instruction-tuned Llama-3.1-70B-Instruct model \citep{llama}, is guided via prompting (see \Cref{app:prompts} for the used prompts).

\paragraph{I. Generation of Events.} The initial step generates a set of event descriptions, which act as instances to be interpreted in a real-world scenario.
We ensure diversity by adopting the attribute-based strategy proposed by \citet{NEURIPS2023_ae9500c4}, leading to a balanced distribution of topics, as mentioned in the prompt (Step I in \Cref{tab:examplesg}; topics: \Cref{app:eventtypes}).
We perform few-shot prompting with ten examples.

\paragraph{II. Generation of Backstories.}
Given an event description $s_5$, the second step creates the corresponding backstories $\mathbb{B}(s_5)$.
We specify the emotion the resulting event chain should evoke in the prompt, i.e., for every $s_5$ from the generated set of events, we include $e(b_i \cdot s_5)$ for every $ b_i \in \mathbb{B}(s_5)$ with $i \compacteq 1, \ldots, n$.
Guiding principles are this specific influence on the evoked emotion and narrative coherence.
We generate a different backstory for each emotion category based on a given event.
To address this challenge within a forward-generation LLM, we convert the backward generation task into a forward generation approach by embedding specific instructions into the prompt.

We compare three approaches (\Cref{fig:overview}):
In the Baseline, all requirements are specified within a single prompt (Step II in \Cref{tab:examplesg}).
In our Plan--Construct (PC) chain-of-thought method, we perform story generation decomposition \citep[inspired by][]{clark2022working} and story planning \citep[following][]{razumovskaia-etal-2024-little}.
The method Plan--Construct--Revise (PCR) subsequently revises the chain in consideration of the two previous outputs.

\subsection{Evaluation Methods}
We aim at understanding if including the generated contexts allows humans and automatic classification to estimate the associated emotion more consistently.
At the same time, the created narratives shall be coherent.

\paragraph{Coherence.}
We evaluate coherence with a zero-shot shuffle test \citep{laban-etal-2021-transformer} using an LLM to compute the sequence likelihood of event chains (\Cref{sec:csa}).
The resulting scores are compared against various permutations in which the order of events is shuffled.
A high coherence score (total range from 0 to 1) is assigned to the original event chain if it ranks favorably among its shuffled permutations.

\paragraph{Human Annotation.}
We analyze generated events as well as the impact of backstories on evoked emotions by conducting human annotation studies.
We ask crowd workers to annotate emotions as well as to assess the quality of a sample (3~annotators per instance, see \Cref{app:anno} for the setup).\footnote{We use Prolific (\url{https://www.prolific.com/}).}
Given an event chain (or just one event description), annotators have to select the most prominent emotion the author of the text presumably felt in the end.
Events are additionally annotated on a Likert scale from 1 to 5 regarding vagueness and plausibility, as well as whether they are assumed to be written by a human or by an AI.
An important aspect in the evaluation is that we do not only evaluate the average agreement, but particularly aim to identify the cases where a reasonable context allows for an improved annotation.

For the human annotation, we sample a set of 10 events from our dataset (one from each event type category) and collect 39 annotations for this sample.\footnote{We collect 39 annotations for each event since we intend to compare these to the annotations of 13 corresponding event chains (with 3 annotations each).}
Additionally, we sample 90 more events (nine from each event type category) and collect 3 annotations each, so we can evaluate the quality of the generated texts on a sample of 100 events (10\% of our dataset).
To annotate entire narratives, we collect 780 annotations of event chains, which corresponds to three annotations per 13 backstories, generated by each of the two methods, for each of the 10 sampled events.

\paragraph{Emotion Analysis.}
To assess the impact of additional context on automatic classification, we zero-shot classify emotions.\footnote{We use Llama-3.1-70B-Instruct \citep{llama}.}
For each instance (events, backstories, or event chains), we prompt the model with a set of input templates where each includes a specific emotion label (as shown in \Cref{tab:zsea}) and record emotion category probabilities based on the model's sequence likelihoods.
Therefore, we compute $p(e|\{s_1, ..., s_m\})$ for $m\compacteq1, \ldots, 5$ and every $e \in \mathbb{E}$.
The emotion classification performance is on par with the scores reported by \citep{troiano2023dimensional} on a comparable corpus with the same emotion categories (our approach achieves .54 F1, vs.\ .59). Their work employs a fine-tuned RoBERTa-large model \citep{zhuang-etal-2021-robustly,roberta}.

\begin{table}
	\centering\small\sffamily
	\renewcommand{\arraystretch}{1.25}
	\begin{tabularx}{\linewidth}{lX}
		\toprule
		MT&Prompt Message Template\\
		\cmidrule(r){1-1}\cmidrule(l){2-2}
		{System}&
		You are an expert in emotion analysis on event descriptions. \\ 
		
		{User}&
		A person describes their experience as follows:\newline
		\{text\_instance\}\newline
		What emotion was evoked in the person at the end? As your response, provide only one label from the emotion set: anger, disgust, fear, guilt, joy, sadness, shame, pride, boredom, surprise, trust, relief, no-emotion. \\
		
		{Assistant}&
		\{emotion\} \\
		\bottomrule
	\end{tabularx} 
	\caption{Template of prompt messages for our zero-shot emotion analysis on events and event chains. `MT' refers to the prompt message type as specified in the input for the instruction-tuned LLM.}\label{tab:zsea}
\end{table}

\section{Analysis}\label{sec:data}
Our dataset is based on 1000 generated event descriptions consisting of 15.4 tokens on average (\Cref{app:stats}).
The instances exhibit a low level of vagueness and are plausible (based on human annotation of 100 sampled event descriptions, vagueness: mean 2.50, $\sigma \compacteq 1.20$, plausibility: mean 4.32, $\sigma \compacteq 0.89$).
The relatively high standard deviation for vagueness aligns well with our goal of creating a diverse dataset.
The annotators tend to attribute authorship to humans (mean 3.50, $\sigma \compacteq 0.99$) over AI (mean~2.67, $\sigma \compacteq 1.05$).

We generate 13 backstories for each event, corresponding to the set of emotion categories studied.
Consequently, our dataset comprises 13,000 event chains for each of the three methods.
The Baseline generates backstories averaging 69.8 tokens, while PC backstories average 49.2 tokens and PCR backstories average 79.2 tokens (\Cref{app:stats}).
Examples are shown in \Cref{tab:examples}.
We conduct an analysis for lexical diversity, coherence and emotions to answer our research questions in the following subsections.

\subsection{\RQone}\label{sec:coh}
\newcommand{\gd}[1]{\medmuskip=0mu\relax\textcolor{OliveGreen}{\scriptsize $+#1$}}
\newcommand{\bd}{\medmuskip=0mu\relax\scriptsize $\pm.00$}
\newcommand{\rd}[1]{\medmuskip=0mu\relax\textcolor{Maroon}{\scriptsize $-#1$}}
\begin{table}
	\centering\small
	\begin{tabular}{p{2mm} l r r@{\hspace{.5\tabcolsep}}l r@{\hspace{.5\tabcolsep}}l}
		\toprule
		&&\multicolumn{5}{c}{Mean Coherence of Chains}\\
		\cmidrule(){3-7}
		&&Baseline&\multicolumn{2}{c}{PC}&\multicolumn{2}{c}{PCR}\\
		\cmidrule(r){3-3} \cmidrule(rl){4-5} \cmidrule(l){6-7}
		\multirow{13}{*}{\rotatebox{90}{Emotion-Specific Subsets}}
		&Anger & .75& .78&\gd{.03}& .84&\gd{.09}\\
		&Boredom & .73 & .76 &\gd{.03}& .84&\gd{.11}\\
		&Disgust & .72 & .77 &\gd{.05}& .84&\gd{.12}\\
		&Fear & .67 & .72 &\gd{.05}& .80&\gd{.13}\\
		&Guilt & .77& .79 &\gd{.02}& .86&\gd{.09}\\
		&Joy & .76 & .78 &\gd{.02}& .85&\gd{.09}\\
		&Pride & .81& .78 &\rd{.03}& .85&\gd{.04}\\
		&Relief & .81& .78 &\rd{.03}& .85&\gd{.04}\\
		&Sadness & .76 & .79 &\gd{.03}& .84&\gd{.08}\\
		&Shame & .78 & .79 &\gd{.01}& .86&\gd{.06}\\
		&Surprise & .76 & .79 &\gd{.03}& .83&\gd{.07}\\
		&Trust & .78& .80 &\gd{.02}& .85&\gd{.07}\\
		&No-Emotion & .69& .74 &\gd{.05}& .82&\gd{.13}\\
		\cmidrule(r){1-2}\cmidrule(rl){3-3} \cmidrule(rl){4-5} \cmidrule(l){6-7}
		\multicolumn{2}{l}{Overall Dataset}&  .75&.77&\gd{.02}&.84&\gd{.09}\\[-1.25mm]
		&& {\scriptsize $\sigma=.26$}&\multicolumn{2}{l}{\scriptsize $\sigma=.25$}&\multicolumn{2}{l}{\scriptsize $\sigma=.21$}\\
		\bottomrule
	\end{tabular}
	\caption{Mean coherence scores for event chains in the datasets generated with different methods. Delta values (marked with {\textcolor{OliveGreen}{\scriptsize $+$}}/{\textcolor{Maroon}{\scriptsize$-$}}) show the difference in comparison to the Baseline.}
	\label{tab:csstats}
\end{table}

A requirement for the analysis of evoked emotions is that the data is of high quality.
We first evaluate the coherence scores of the event chains generated by our three methods. 
The results displayed in \Cref{tab:csstats} show that PCR (.84) and PC (.77) both outperform the baseline method (.75):
Story planning does contribute to more coherent narratives.
In a qualitative analysis we further find that PC and PCR generate more diverse backstories than the baseline (see \Cref{app:examples}).
The variation in story coherence across different emotion categories is relatively low.  
All scores are on an acceptable level and the content diversity of the generated texts is high (see \Cref{sec:res}).
Fear shows the lowest coherence scores for all methods.

\begin{figure}[]
	\centering
	\begin{tikzpicture}
		\begin{axis}[
			xlabel={Number of Best Chains Selected per Event},
			ylabel={Fleiss' Kappa},
			ymin=0, ymax=1,
			xmin=1, xmax=13,
			xtick={1,2,...,13},
			tick align=outside,
			tick pos=left,
			width=\linewidth,
			height=5.5cm,
			grid=both,
			legend style={align=left, anchor=north east},
			legend cell align={left},
			tick label style={font=\small},
			legend style={font=\small},
			font=\small
			]
			\addplot[black, thick] coordinates {(1, 0.240)(13, 0.240)};
			\addlegendentry{Events}
			
			\addplot[baselinecolor, thick, mark=triangle] coordinates {(1, 0.675) (2, 0.640) (3, 0.562) (4, 0.485) (5, 0.430) (6, 0.380) (7, 0.338) (8, 0.288) (9, 0.253) (10, 0.222) (11, 0.194) (12, 0.170) (13, 0.147)};
			\addlegendentry{Baseline Chains}
			
			\addplot[pcrcolor, thick, mark=o] coordinates {(1, 0.748) (2, 0.649) (3, 0.538) (4, 0.494) (5, 0.459) (6, 0.425) (7, 0.390) (8, 0.349) (9, 0.309) (10, 0.274) (11, 0.243) (12, 0.213) (13, 0.188)};
			\addlegendentry{PCR Chains}
		\end{axis}
	\end{tikzpicture}
	\caption{Emotion annotation agreement on events in comparison to Baseline and PCR chains.}
	\label{fig:humanaggplot}
\end{figure}
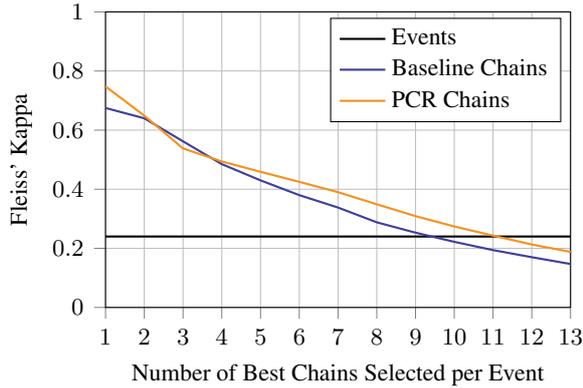

\paragraph{Potential Emotion Leakage in Backstories.}
The generated backstories are designed to avoid explicit references to the targeted emotions.
To assess adherence to this aim, we analyze potential emotion leakage.
We use ChatGPT \citep{chatgpt} to compile lists of 25 synonyms for each of our emotion categories and manually refine these to only contain unique terms.
We count cases where any of these terms or the emotion word itself appear in the backstories as emotion leakage.

We find only a low issue with leakage rates in backstories: 2\% of the background stories contain an emotion word in the Baseline (310 out of 13,000), 1\% for PC (165 out of 13,000), and 5\% for PCR (682 out of 13,000). 
Paraphrasing these terms or filtering out the instances could mitigate this issue.
However, we chose to retain the few instances of emotion leakage in our analysis to maintain a straightforward evaluation process, recognizing that filtering might inadvertently also introduce errors.
This approach highlights opportunities for refinement in future research.

\subsection{\RQtwo}
We now assess the impact of our generated contexts on the clarity of emotion interpretations based on results of the human annotation on a sample of our dataset.
This sample consists of 10 distinct events, each enriched by one Baseline backstory as well as one PCR backstory for each of the 13 emotion categories studied, i.e.\ two sets of 130 event chains.

Our goal is to determine whether the introduction of the contexts clarifies the emotion categorization given the event chains for humans compared to when assessing the individual events.
Therefore, we calculate the inter-annotator agreement (Fleiss' kappa), which is .24 on events,  .15 for Baseline chains and .19 for PCR chains. 
However, we should take into account that this evaluation encompasses the backstories generated across all the different emotion categories for each event.
Since an event may not evoke all possible emotions even when viewed in context, it is reasonable to expect that disambiguation would only be feasible for certain relevant emotion categories.
Therefore, we evaluate if there are valuable contexts available to the annotators by analyzing only a selection of the generated chains relevant to each event. 

\begin{table}
	\centering\small \setlength{\tabcolsep}{2pt}
	\begin{tabular}{l@{\hspace{.75\tabcolsep}}rrrrrrrrrrrrr}
		\toprule
		\diagbox[innerleftsep=0cm,width=1.25cm,height=1.4cm]{$e(b{\mkern1.5mu{\cdot}\mkern1.5mu}s_5)$}{$e_A$} & \rotatebox[origin=c]{90}{Anger}&\rotatebox[origin=c]{90}{Boredom}&\rotatebox[origin=c]{90}{Disgust}&\rotatebox[origin=c]{90}{Fear}&\rotatebox[origin=c]{90}{Guilt}&\rotatebox[origin=c]{90}{Joy}&\rotatebox[origin=c]{90}{Pride}&\rotatebox[origin=c]{90}{Relief}&\rotatebox[origin=c]{90}{Sadness}&\rotatebox[origin=c]{90}{Shame}&\rotatebox[origin=c]{90}{Surprise}&\rotatebox[origin=c]{90}{Trust}&\rotatebox[origin=c]{90}{No-Emot.}\\
		\cmidrule(r{.2em}){1-1}\cmidrule(r{.2em}l{.2em}){2-2}\cmidrule(r{.2em}l{.2em}){3-3}\cmidrule(r{.2em}l{.2em}){4-4}\cmidrule(r{.2em}l{.2em}){5-5}\cmidrule(r{.2em}l{.2em}){6-6}\cmidrule(r{.2em}l{.2em}){7-7}\cmidrule(r{.2em}l{.2em}){8-8}\cmidrule(r{.2em}l{.2em}){9-9}\cmidrule(r{.2em}l{.2em}){10-10}\cmidrule(r{.2em}l{.2em}){11-11}\cmidrule(r{.2em}l{.2em}){12-12}\cmidrule(r{.2em}l{.2em}){13-13}\cmidrule(l{.2em}){14-14}
		Anger& \cellcolor{blue!16}5& \cellcolor{blue!0}0& \cellcolor{blue!0}0& \cellcolor{blue!16}5& \cellcolor{blue!0}0& \cellcolor{blue!0}0& \cellcolor{blue!0}0& \cellcolor{blue!10}3& \cellcolor{blue!36}11& \cellcolor{blue!3}1& \cellcolor{blue!10}3& \cellcolor{blue!0}0& \cellcolor{blue!6}2\\
		Boredom& \cellcolor{blue!16}5& \cellcolor{blue!0}0& \cellcolor{blue!6}2& \cellcolor{blue!3}1& \cellcolor{blue!0}0& \cellcolor{blue!6}2& \cellcolor{blue!10}3& \cellcolor{blue!0}0& \cellcolor{blue!26}8& \cellcolor{blue!0}0& \cellcolor{blue!3}1& \cellcolor{blue!0}0& \cellcolor{blue!26}8\\
		Disgust& \cellcolor{blue!10}3& \cellcolor{blue!3}1& \cellcolor{blue!20}6& \cellcolor{blue!3}1& \cellcolor{blue!0}0& \cellcolor{blue!13}4& \cellcolor{blue!0}0& \cellcolor{blue!3}1& \cellcolor{blue!6}2& \cellcolor{blue!6}2& \cellcolor{blue!10}3& \cellcolor{blue!3}1& \cellcolor{blue!20}6\\
		Fear& \cellcolor{blue!13}4& \cellcolor{blue!0}0& \cellcolor{blue!3}1& \cellcolor{blue!40}12& \cellcolor{blue!6}2& \cellcolor{blue!0}0& \cellcolor{blue!0}0& \cellcolor{blue!3}1& \cellcolor{blue!3}1& \cellcolor{blue!6}2& \cellcolor{blue!13}4& \cellcolor{blue!0}0& \cellcolor{blue!10}3\\
		Guilt& \cellcolor{blue!10}3& \cellcolor{blue!0}0& \cellcolor{blue!0}0& \cellcolor{blue!10}3& \cellcolor{blue!20}6& \cellcolor{blue!0}0& \cellcolor{blue!6}2& \cellcolor{blue!13}4& \cellcolor{blue!13}4& \cellcolor{blue!10}3& \cellcolor{blue!3}1& \cellcolor{blue!0}0& \cellcolor{blue!13}4\\
		Joy& \cellcolor{blue!3}1& \cellcolor{blue!0}0& \cellcolor{blue!3}1& \cellcolor{blue!0}0& \cellcolor{blue!0}0& \cellcolor{blue!6}2& \cellcolor{blue!23}7& \cellcolor{blue!23}7& \cellcolor{blue!16}5& \cellcolor{blue!3}1& \cellcolor{blue!3}1& \cellcolor{blue!3}1& \cellcolor{blue!13}4\\
		Pride& \cellcolor{blue!10}3& \cellcolor{blue!0}0& \cellcolor{blue!0}0& \cellcolor{blue!0}0& \cellcolor{blue!10}3& \cellcolor{blue!0}0& \cellcolor{blue!33}10& \cellcolor{blue!23}7& \cellcolor{blue!0}0& \cellcolor{blue!0}0& \cellcolor{blue!0}0& \cellcolor{blue!3}1& \cellcolor{blue!20}6\\
		Relief& \cellcolor{blue!10}3& \cellcolor{blue!0}0& \cellcolor{blue!0}0& \cellcolor{blue!16}5& \cellcolor{blue!0}0& \cellcolor{blue!0}0& \cellcolor{blue!3}1& \cellcolor{blue!20}6& \cellcolor{blue!13}4& \cellcolor{blue!0}0& \cellcolor{blue!10}3& \cellcolor{blue!6}2& \cellcolor{blue!20}6\\
		Sadness& \cellcolor{blue!6}2& \cellcolor{blue!0}0& \cellcolor{blue!0}0& \cellcolor{blue!10}3& \cellcolor{blue!0}0& \cellcolor{blue!3}1& \cellcolor{blue!3}1& \cellcolor{blue!3}1& \cellcolor{blue!50}15& \cellcolor{blue!0}0& \cellcolor{blue!10}3& \cellcolor{blue!3}1& \cellcolor{blue!10}3\\
		Shame& \cellcolor{blue!0}0& \cellcolor{blue!3}1& \cellcolor{blue!0}0& \cellcolor{blue!10}3& \cellcolor{blue!0}0& \cellcolor{blue!10}3& \cellcolor{blue!6}2& \cellcolor{blue!0}0& \cellcolor{blue!3}1& \cellcolor{blue!36}11& \cellcolor{blue!10}3& \cellcolor{blue!0}0& \cellcolor{blue!20}6\\
		Surprise& \cellcolor{blue!13}4& \cellcolor{blue!0}0& \cellcolor{blue!3}1& \cellcolor{blue!0}0& \cellcolor{blue!3}1& \cellcolor{blue!10}3& \cellcolor{blue!6}2& \cellcolor{blue!0}0& \cellcolor{blue!23}7& \cellcolor{blue!3}1& \cellcolor{blue!20}6& \cellcolor{blue!6}2& \cellcolor{blue!10}3\\
		Trust& \cellcolor{blue!10}3& \cellcolor{blue!3}1& \cellcolor{blue!0}0& \cellcolor{blue!0}0& \cellcolor{blue!0}0& \cellcolor{blue!3}1& \cellcolor{blue!23}7& \cellcolor{blue!30}9& \cellcolor{blue!6}2& \cellcolor{blue!0}0& \cellcolor{blue!0}0& \cellcolor{blue!6}2& \cellcolor{blue!16}5\\
		No-Emot.& \cellcolor{blue!20}6& \cellcolor{blue!0}0& \cellcolor{blue!0}0& \cellcolor{blue!6}2& \cellcolor{blue!0}0& \cellcolor{blue!6}2& \cellcolor{blue!6}2& \cellcolor{blue!26}8& \cellcolor{blue!6}2& \cellcolor{blue!3}1& \cellcolor{blue!6}2& \cellcolor{blue!0}0& \cellcolor{blue!16}5\\
		$\Sigma$&42&3&11&35&12&18&37&47&62&22&30&10&61\\
		\bottomrule
	\end{tabular}
	\caption{Prompted emotion $e(b{\mkern1.5mu{\cdot}\mkern1.5mu}s_5)$ vs.\ annotated emotion $e_a$ on 130 PCR event chains (3 annotations each), 10 per $e \in e(b{\mkern1.5mu{\cdot}\mkern1.5mu}s_5)$.}
	\label{tab:pe_annoe_m_igc}
\end{table}

We find that (see \Cref{fig:humanaggplot}) for PCR chains, a vast majority -- 10 out of the 13 backstories generated for each event -- successfully provide informative context that enhances clarity in emotion analysis for annotators.
Our approach effectively disambiguates emotion interpretations of given events.

To further understand how humans interpret emotions within the contextualized events provided in our dataset, we investigate the specific emotion categories assigned during the annotation process.
In \Cref{tab:pe_annoe_m_igc}, the counts of annotated labels (columns) of the PCR chains generated for the different prompted emotions (rows) are organized.
Annotators recognize the prompted emotion most often for fear, pride, sadness, and shame.
Notably, certain patterns arise: Chains designed to trigger anger are often annotated as sadness, while those prompted for trust are frequently classified as relief.
Additional analysis confirms that such typical overlaps among the annotated emotion categories are evident in our data (see \Cref{app:anno}).
Overall, for most emotions the prompted emotion is the category which is most often annotated by the human annotators.
This further reinforces the efficacy of our context generation framework in clarifying emotion interpretations for humans.

\subsection{\RQthree}

\paragraph{Annotation Correlation.}
To understand if our automatic emotion analysis aligns with human annotations, we compute the Spearman correlation between the probabilistic prediction score and the proportion of annotators who identified the respective emotion.
The overall results reveal a correlation of .47 for events and .37 for chains -- both significant (p < .001).
Further analysis of the correlation shows that events exhibit greater fluctuations across emotion categories, whereas assessments of chains are more consistent (see \Cref{app:corrclf}).

\paragraph{The Overall Influence of Backstories in Emotion Analysis.}
The probability of each emotion category being evoked on average across the 1000 events (see \Cref{tab:3levelemo}, under the column labeled E) reveals a typical distribution of emotions associated with general descriptions of human experiences.
It favors frequently occurring categories, such as joy, surprise, and no emotion.
In comparison, the average scores for the emotions evoked given entire chains (columns labeled C) generated by the three methods is .24, .19 and .27 respectively.
This is considerably higher than the average probability of the emotion on events (.08).
The standard deviation shows substantial variability which indicates that there are several cases where the intended emotion has a very high probability.
The generated context does disambiguate emotion categorization for our automatic system.

\begin{table}
	\setlength{\tabcolsep}{2.75pt}
	\centering\small
	\begin{tabular}{p{2mm} l r rr rr@{\hspace{.25\tabcolsep}}l rr@{\hspace{.25\tabcolsep}}l}
		\toprule
		&&&\multicolumn{2}{c}{Baseline}&\multicolumn{3}{c}{PC}&\multicolumn{3}{c}{PCR}\\
		\cmidrule(r{.4em}){4-5} \cmidrule(r{.4em}l{.4em}){6-8} \cmidrule(l{.4em}){9-11}
		&&E& B&C & B&\multicolumn{2}{c}{C}& B&\multicolumn{2}{c}{C}\\
		\cmidrule(r{.4em}){3-3} \cmidrule(r{.4em}l{.4em}){4-4} \cmidrule(r{.4em}l{.4em}){5-5} \cmidrule(r{.4em}l{.4em}){6-6} \cmidrule(r{.4em}l{.4em}){7-8} \cmidrule(r{.4em}l{.4em}){9-9} \cmidrule(r{.4em}l{.4em}){10-11}
		\multirow{13}{*}{\rotatebox{90}{Emotion-Specific Subsets}}
		&Anger      & .02& .36&  .27&  .25&  .17& \rd{.10}& .32& .24& \rd{.03}\\
		&Boredom& .01& .09&  .07&  .08&  .07& \bd{}& .14&  .12& \gd{.05}\\
		&Disgust   & .00& .22&  .16&  .25&  .17& \gd{.01}& .30&  .24& \gd{.08}\\
		&Fear         & .01& .46&  .28&  .30&  .15& \rd{.13}& .53& .28& \bd{}\\
		&Guilt        & .00& .31&  .24&  .24&  .16& \rd{.08}& .34&  .24& \bd{}\\
		&Joy          & .33& .11& .24&  .12&  .26& \gd{.02}& .15&  .27& \gd{.03}\\
		&Pride       & .01& .23&  .23&  .16&  .15& \rd{.08}& .35&  .24& \gd{.01}\\
		&Relief      & .08& .33& .56&  .35&  .53& \rd{.03}& .32& .61& \gd{.05}\\
		&Sadness& .01& .37&  .31&  .32&  .23& \rd{.08}& .50&  .42& \gd{.11}\\
		&Shame   & .00& .28& .23&  .20&  .16& \rd{.07}& .32&  .27& \gd{.04}\\
		&Surprise & .14& .19&  .37&  .14&  .24& \rd{.13}& .15&  .30& \rd{.07}\\
		&Trust       & .02& .08& .07&  .11&  .08& \gd{.01}& .19& .13& \gd{.06}\\
		&No-Emot. & .37& .02& .06& .03&.06&\bd{}& .04&.09&\gd{.03}\\
		\cmidrule(r{.4em}){1-2}\cmidrule(r{.4em}l{.4em}){3-3} \cmidrule(r{.4em}l{.4em}){4-5} \cmidrule(r{.4em}l{.4em}){6-8} \cmidrule(l{.4em}){9-11}
		\multicolumn{2}{l}{Overall}&.08 & .23&.24 & .20&.19&\rd{.05} & .28&.27&\gd{.03} \\[-1.mm]
		\multicolumn{2}{r}{\scriptsize $\sigma$}& {\scriptsize $.24$}&{\scriptsize $.38$}&{\scriptsize $.37$}&{\scriptsize $.36$}&\multicolumn{2}{l}{\scriptsize $.34$}&{\scriptsize $.41$}&\multicolumn{2}{l}{\scriptsize $.38$}\\
		\bottomrule
	\end{tabular}
	\caption{Averaged results for probabilistic emotion analysis given events (E) or, as generated by different methods for prompted emotions, backstories (B) or entire event chains (C). Delta values (marked with {\textcolor{OliveGreen}{\scriptsize $+$}}/{\scriptsize$\pm$}/{\textcolor{Maroon}{\scriptsize$-$}}) show the difference in comparison to the Baseline.}
	\label{tab:3levelemo}
\end{table}

The chains generated through PCR evoke the desired emotions more effectively than those produced with the baseline method, as confirmed by human annotations (see \Cref{fig:humanaggplot}).
PCR is particularly successful in generating event chains for the emotions of relief and sadness.  
However, capturing emotions such as boredom or trust proves to be challenging, possibly due to the insufficient information in the context or backstory, or because these emotions require very specific events to be effectively evoked.
Joy appears to be less distinctly recognized in contextualized events compared to isolated ones, suggesting that this emotion category does not benefit from our approach.
Further analysis of discrete predicted labels reveals common misclassifications, particularly between guilt and shame, as well as between disgust and anger (see \Cref{app:pred}).
Moving forward, generating a larger initial sample size of target events, along with a final filtering process for unsuccessful cases, could produce a large number of narratives where the intended emotion responses are evoked for these particular categories.

\medskip\noindent%
In summary, our analysis shows that systems recognize modifications to the emotion interpretation of events under the inclusion of contextual backstories.
Nonetheless, certain categories present challenges for clear emotion evocation through contexts, necessitating further exploration.

\paragraph{Investigating the Specific Influence of Backstories in Emotion Analysis.} 
We now aim to understand how the generated backstories facilitate the observed modifications.
Therefore, we investigate the emotion analysis derived solely based on the backstories, i.e.\ event chains without the target event (see \Cref{tab:3levelemo}, under the columns labeled B).
We compare these values to the values for entire chains (columns labeled C) to measure how specifically the incorporation of the target event alters the emotion interpretation. 

For certain emotions, such as relief and surprise, the PCR backstories alone show a substantially lower probability of evoking the specific emotion compared to the entire event chain.
In contrast, for emotions like fear and pride, the PCR backstories have a higher probability of eliciting the intended emotion than the complete chain.
These distinct patterns show that the manner of emotion evocation through generated backstories varies across different categories.
While some emotions can be effectively elicited through the contextualization of the target event, others are predominantly grounded in the narrative of the backstory.

\begin{figure}[]
	\centering
	\begin{tikzpicture}
		\begin{axis}[width=\linewidth,
			height=5.5cm,
			xtick={1,2,3,4,5},
			ytick={0,0.1,...,1.0},
			yticklabels={0, 0.1, 0.2, 0.3, 0.4, 0.5, 0.6, 0.7, 0.8, 0.9, 1},
			tick align=outside,
			tick pos=left,
			ymin=0,
			ymax=1,
			xmin=0.75,
			xmax=5.25,
			grid=major,
			domain=1:5,
			legend pos=north west, tick label style={font=\small},
			legend style={font=\small},ylabel={$p(e(b{\mkern1.5mu{\cdot}\mkern1.5mu}s_5)|\{s_1, ..., s_m\})$},
			legend cell align={left},
			xlabel={Number of first sentences of chains $= m$},
			font=\small]
			\addplot[thick, fearcolor, mark=*] coordinates { (1,  0.355) (2,  0.403) (3,  0.429) (4,  0.532) (5,  0.281) };
			\addlegendentry{Fear Chains}
			\addplot[thick, pridecolor, mark=x] coordinates { (1,  0.236) (2,  0.263) (3,  0.295) (4,  0.353) (5,  0.243) };
			\addlegendentry{Pride Chains}
			\addplot[thick, reliefcolor, mark=o] coordinates { (1,  0.072) (2,  0.140) (3,  0.208) (4,  0.324) (5,  0.612) };
			\addlegendentry{Relief Chains}
			\addplot[thick, surprisecolor, mark=triangle] coordinates { (1,  0.164) (2,  0.116) (3,  0.119) (4,  0.152) (5,  0.304) };
			\addlegendentry{Surprise Chains}
		\end{axis}
	\end{tikzpicture}
	\caption{Mean of trajectories of predicted emotion probability scores in event chains for selected emotions ($\{s_1, ..., s_5\}$, generation method: PCR).}
	\label{fig:emotraj}
\end{figure}

To explore this further, we investigate the trajectories of emotion prediction throughout the narratives, analyzing how the emotion scores evolve from sentence to sentence (see \Cref{fig:emotraj}).
These plots not only show the probabilities associated with backstories (represented by $m=4$ in the plots) and entire event chains ($m=5$) but also probabilities derived from preceding segments of the narrative chain (indicated by $m\leq3$ values).
The plots illustrate the emotion trajectory for the four prominent categories previously discussed.
More detailed visualizations of all emotions are given in \Cref{app:emotraj}.

For the emotions fear and pride, we observe a steady increase followed by a decline in probability upon the introduction of the target event.
This decline can be explained by our earlier findings, which suggest that the generated events ($s_5$) are not particularly aligned with these emotions, indicating that these have to derive from the context established in the backstory.\footnote{We chose to retain these instances in our analysis to evaluate the effectiveness of our method across any events and emotions. However, for future applications of the dataset, it may be advisable to filter these cases.}
Conversely, the emotion relief exhibits a steady increase in probability throughout the narrative, with the most significant surge occurring upon the introduction of the final event.
This pattern aligns with Ekman’s concept of ``programs'' \citep[as further described by][]{troiano2023dimensional}, as such complex emotions develop through a progression, which is realized in our narratives.
\Cref{tab:exrm} shows this on examples generated by PCR for the category relief: the five stories with the highest emotion delta, i.e.\ score increase from backstory to the full chain in emotion analysis.

\begin{table*}[]
	\centering\small\sffamily
	\renewcommand{\arraystretch}{1.4}
	\begin{tabularx}{\linewidth}{@{\hspace{1.1mm}}X@{\hspace{1.1mm}}}
		\toprule
		I lost my grandfather's calculator, a gift I had treasured for years. The teacher announced a crucial math problems exercise, one that would determine our grades for the semester. I rummaged through my backpack and desk, but couldn't find a spare calculator. The teacher began explaining the exercise, and I realized I had no way to participate without a calculator. \textit{The teacher passed around a calculator for us to use during the math problems exercise.}\\
		I received an urgent work assignment with a tight deadline that required my undivided attention. My roommate threw an impromptu party in the living room, immediately filling the air with loud music and chatter. The music grew louder, making it increasingly difficult for me to concentrate on my task. The party reached its peak, with the music blasting at an ear-splitting decibel that made it impossible for me to focus. \textit{The music suddenly stopped when someone accidentally knocked over the bluetooth speaker.}\\
		My friend pulled out his pocket knife and started flipping it open and closed near the campfire, narrowly missing his own fingers. As he continued to play with the knife, he began to twirl it around his fingers, coming close to slashing his own arm. I asked him to stop, but he ignored me, using the knife to cut a branch that snapped back and almost hit me. The tension between us grew as he continued to handle the knife carelessly, causing me to flinch every time he came near me with it. \textit{As we were setting up camp near the lake, my friend accidentally dropped his pocket knife into the water.}\\
		I stopped to take a photo of the waterfall and lost sight of the group in the dense foliage. I walked a short distance down the trail, expecting to catch up with the group, but they were nowhere in sight. As I turned back to retrace my steps, I realized I had taken a wrong turn and was now facing an unfamiliar section of the trail. The sound of the waterfall grew fainter, and I was surrounded by an unsettling silence, with no sign of the group anywhere. \textit{Our guide blew the rescue whistle to signal for us to regroup after we got separated while hiking near the waterfall.}\\
		I got lost in the city while trying to find a restaurant for lunch, unable to read the signs or ask for help due to the language barrier. I tried to ask multiple locals for directions, but none spoke English, and I was forced to rely on hand gestures and guesswork. I finally found my way back to the hotel, late for a group meeting, where my fellow travelers were worried about me and struggled to communicate with the hotel staff themselves. I received a reminder about my upcoming meeting with a local business partner, who I knew didn't speak English, and I began to worry about how I would negotiate a crucial business deal. \textit{The tour guide handed out translation devices to help us navigate the language barrier during our trip abroad.}\\
		\bottomrule
	\end{tabularx}
	\caption{Examples of relief stories with highest emotion delta (score increase from backstory to full chain).}
	\label{tab:exrm}
\end{table*}

Our analysis shows that generated backstories can influence the emotion analysis of events in different ways.
It is important to note that in certain cases, evoking emotions mainly from the backstory is not an inherent drawback.
In fact, this approach can be essential for eliciting complex emotions that would otherwise remain unexplored.
Our data contain events with ambiguous emotions as well as events devoid of specific emotion, which could potentially be emotionally charged through contextualization.
An investigation of our data shows that, if the model shows low uncertainty in emotion analysis for events, this indicates that these events are easier to disambiguate using coherent narratives (\Cref{app:cohcorr}).
Further distinction between such types of events justifies additional analysis and is a starting point for future work.

In summary, our analyses underscore the complex interplay between context and emotion evocation, highlighting that the foundation established by a narrative backstory can be critical for the perception and experience of particular emotions, while also revealing distinctive pathways for how emotions manifest within narratives.

\section{Conclusion}\label{sec:conc}
We explore the relationship between contextual backstories and emotion interpretations in event descriptions, in cases in which events alone do not provide sufficient context for a non-ambiguous interpretation.
We therefore make potential interpretations by readers explicit.
Our paper contributes a dataset to study such additional context and proposes a set of methods, inspired by story planning, to achieve coherent and diverse narratives.
Indeed, our additional context does increase inter-annotator agreement and is correctly recognized by automatic prediction systems.
Specifically, our approach substantially enhances the evocation of emotions such as relief and sadness, while emotions like boredom and trust are less effectively evoked with the additional context we provide.
A more comprehensive analysis using further emotion-labeled story datasets is a valuable future task to enhance our findings.
Concerning the language-dependence of our work, \citet{schafer-etal-2025-localization} have demonstrated similar results for German narratives; however, their findings suggest that the generated data is less effective in eliciting desired emotions in that language.
This discrepancy likely arises because the large language model is more proficient in English.

Next to the modeling and dataset perspective, our method can also be seen as a contribution to Explainable Artificial Intelligence (XAI).
By generating contextual backstories, our approach can be used to clarify uncertainties and probabilities in automatic classifications.
This allows model decisions to be communicated more clearly and alternative interpretations to be demonstrated.

Our dataset \textsc{EBS} (\textsc{E}motional \textsc{B}ack\textsc{S}tories) can be used to explore various dimensions, such as emotion changes in event chains or the impact of demographic differences on contextualized emotion interpretations.
For example, \citet{Schaefer2026} show how it can be utilized to examine how appraisals evolve in the context of emotions, particularly by investigating the trajectories of appraisal scores within narratives. Consequently, the dataset offers further understanding of how emotions can be modeled from textual event descriptions.
Furthermore, our dataset can be instrumental in distinguishing between types of events where initially the evoked emotions are unclear -- specifically, those that are devoid of emotion versus those that evoke multiple emotions.
This distinction underscores how contextual backstories can influence emotion responses in particular events and contributes to clarifying scenarios where initial emotion analysis may be unclear.
In our dataset, variations in how backstories modify emotions elicited by events reveal distinct patterns, which can help differentiate between these types of events.
Moreover, our dataset serves as novel training data for models specializing in contextualized emotion analysis.
By encompassing a variety of backstories that elicit different emotion responses from the same event, researchers can use our data to develop and test nuanced classifiers.

While we provide valuable insights and a novel dataset, further research is needed to expand on our findings.
Specifically, future work should investigate how our approach of generating narrative backstories and the resulting dataset can be tailored for various applications, such as improving dialogue systems or enhancing the recognition of human emotions in interactions with language models.
In particular, the potential utility of our approach needs to be further explored in order to develop systems that better interpret the complexity of human emotions in different contexts.

\section*{Acknowledgments}
This work has been supported by the German Research Foundation (DFG) in the project KL2869/1–2 (CEAT, project number 380093645).

\section*{Limitations}
While our study offers valuable insights into emotion analysis through generated narratives, it is important to acknowledge several limitations.  
First, our dataset is based on only 1,000 event descriptions generated by one model, which may limit the diversity and generalizability of our findings across different contexts and emotion categories.  
Incorporating a second, distinct language model to validate the emotion in the generated narratives could enhance robustness by discarding instances with mismatched emotions. 
Employing multiple language models in our analysis would further strengthen the significance of our experimental results.
Second, the human annotation study was conducted solely on a sample of the data, which may not fully capture the complexity of the emotion responses present in the entire dataset.
Additionally, the automatic model used to assess the entire corpus is the same one that generated the data, potentially reinforcing interpretations.
Moreover, this model has not been validated for contextualized emotion analysis, leaving its accuracy uncertain in this context.
To facilitate reproducibility, we offer access to our code, dataset, model predictions, and the human annotations, enabling further exploration and validation of our findings.

\section*{Ethical Considerations}

We relied on crowd workers from Prolific to conduct the human evaluation.
The annotators were paid £9.85--£10.59 per hour.
All participants were shown a consent form containing the information and requirements regarding the study.
They had to confirm their acceptance to be able to participate in the study.
We provided an email address to contact us in case of problems during and after the study.
The total cost of the annotation study including platform fees and taxes amounted to £511.64.

In developing our emotion analysis framework, we are committed to ethical considerations surrounding the use of generated narratives.  
We recognize the potential for biases in emotion interpretation, which may inadvertently reinforce stereotypes or misrepresent certain emotion contexts.  
To mitigate these risks, we emphasize transparency in our methodology and provide access to our datasets and models, allowing others to examine and address ethical implications in their applications.  
Additionally, we advocate for thorough evaluation of the generated narratives to ensure they align with ethical standards and contribute positively to understanding human emotions.
Our work poses a risk of automatic emotion manipulation through the intentional evocation of specific feelings and may also reinforce biases in emotion classification due to the repeated use of automatic systems.

ChatGPT \citep{chatgpt} was used to gain inspiration for formulations of initial notes for some of the text of this paper, as well as to find typos.

\section{Bibliographical References}\label{sec:reference}
\bibliographystyle{lrec2026-natbib}
\bibliography{lit}

\appendix

\section{Event Types and Objects}
\label{app:eventtypes}

\Cref{tab:eto} shows our list of 10 event types along with examples of 20 corresponding objects.
These values serve as attributes in the prompt which initially generates diverse events that we subsequently develop backstories for.

\begin{table*}[t]\centering\scriptsize\sffamily
	\renewcommand{\arraystretch}{1.2}
	\begin{tabularx}{\textwidth}{p{1.75cm}X}
		\toprule
		Event Type& Event Type Objects\\ 
		\cmidrule(r){1-1}\cmidrule(l){2-2}
		Social Gathering&chairs, tables, food platters, drinks, napkins, decorations, music speakers, games, invitation cards, host, guests, tablecloths, candles, balloons, party favors, photo booth, name tags, cutlery, glasses, ice bucket, ... \\
		Educational Activity&textbooks, notebooks, pencils, whiteboard, markers, projector, handouts, calculator, overhead projector, globe, poster board, computer, scissors, gluestick, craft supplies, timers, textual resources, reference books, tables, student desks, ... \\
		Recreational and Nature Activity&hiking boots, backpacks, water bottles, first-aid kit, campfire supplies, nature guide, binoculars, tent, sleeping bags, camping chairs, fishing gear, bicycles, kayaks, picnic basket, coolers, maps, sunscreen, bug spray, fishing rods, swimming gear, ... \\
		Cultural and Community Event&stage, performers, sound system, projector, festival tickets, food stalls, craft booths, cultural displays, artworks, costumes, brochures, community posters, instruments, banners, seating areas, local products, vendors, volunteers, refreshments, cultural symbols, ... \\
		Professional Development&business cards, presentation slides, notebooks, pens, projector, handouts, networking tools, feedback forms, laptops, name badges, workshops, career fair flyers, industry reports, coffee cups, panel discussion guides, training materials, lecture notes, team-building activities, case studies, clipboards, ... \\
		Celebration&cake, candles, party hats, balloons, confetti, party favors, streamers, drinks, gift bags, music playlist, photo booth, decorations, invitation cards, celebration banner, tables, chairs, food platters, glasses, plates, silverware, ... \\
		Artistic Performance&stage, costumes, sets, props, lights, sound equipment, musical instruments, audience seats, backdrops, tickets, makeup kit, rehearsal schedule, choreography notes, great hits collection, piano, amplifiers, performance schedule, music sheets, playbill, actors, ... \\
		Competition&trophies, medals, referee kit, scoreboard, team jerseys, game equipment, whistle, competition schedule, event tickets, player registration, crowd barriers, timing devices, venue maps, registration forms, team banners, score sheets, first aid kits, video cameras, performance analytics, heat sheets, ... \\
		Family and Relationships&family photo albums, toys, family tree chart, gift cards, family recipe book, family calendars, cameraman, outdoor equipment, gifts, personalized items, family game night materials, storybooks, name tags, family bonding games, sentimental objects, blankets, picnic spreads, board games, interactive toys, family outings, ... \\
		Transportation and Travel Event&maps, itineraries, backpacks, suitcases, boarding passes, tickets, travel guides, snacks, passports, travel pillows, sunscreen, water bottles, portable chargers, cameras, compact umbrellas, guidebooks, tour buses, airport shuttles, magazine subscriptions, reservation forms, ... \\
		\bottomrule
	\end{tabularx}
	\caption{List of event types and objects used in prompt templates to create diverse event descriptions.}\label{tab:eto}
\end{table*}

\section{Full Text Prompts}
\label{app:prompts}

\Cref{tab:ftprompts} shows the full text of the prompts we use to generate our data according to the different methods.
The interaction with the model comprises three message types\footnote{\url{https://www.llama.com/docs/model-cards-and-prompt-formats/llama3_1/}}:
a ``system''  message that establishes the context for the interaction and includes general guidelines; a ``user'' message that encapsulates the specific inputs, requirements, and instructions for the task; and an ``assistant''  message that represents the model's response based on the provided context. 

\textbf{Prompt 1} is used to generate the concluding event and aims to ensure a balanced topic variance through the incorporation of specific attributes.
To ensure that the generated event descriptions are concise, we implement few-shot prompting by including ten examples of event descriptions in the prompt.
As values for the different attributes, we use 10 distinct event types, each associated with 100 unique objects.
We acquire these values from several interactions with ChatGPT \citep{chatgpt}.
For example, an event type can be a ``Social Gathering'' with objects like ``balloons''.
The full list of event types with examples of objects is given in \Cref{app:eventtypes}. 

For each event generated in the first step, we aim to create a corresponding backstory comprised of four preceding events, tailored to influence the specific emotion responses of the last event.
In the baseline approach, this process is implemented using a single prompt (\textbf{Prompt 2} as shown in \Cref{tab:ftprompts}).
For the other methods, the process is further broken down into sub-steps:
Story planning using \textbf{Prompt 2.1:} For each final event, the LLM is prompted to first generate a story plan that outlines plausible explanations for the emotion responses tied to the concluding event. This preemptive planning increases coherence and relevance in the backstories.  Based on the crafted story plan, the LLM should generate the sequential backstory comprising four events. This ensures that the events are narratively connected and describe the context leading up to the pivotal final event.
Event chain generation using \textbf{Prompt 2.2:} After generating the event chain, we conduct a summarization step to ensure uniformity and clarity. 
Additionally, our third method uses \textbf{Prompt 2.3} to revise the generated event chain in an attempt to ensure adherence to the defined requirements.

\begin{table*}[t!]
	\centering\sffamily\small
	\renewcommand{\arraystretch}{.75}
	\begin{tabularx}{\linewidth}{p{2mm}p{3mm}X}
		
		\toprule
		\#&MT&Prompt Text\\
		\cmidrule(r){1-1}\cmidrule(rl){2-2}\cmidrule(l){3-3}
		
		\multirow{3}{*}{\rotatebox{90}{Prompt 1}}
		&\multirow{2}{*}{\rotatebox{90}{system}} 
		& You are a person describing an event which you have experienced. 
		
		10 examples of such event descriptions are as follows:  
		0: The phone rang. 
		1: A cat meowed. 
		2: The car engine sputtered to a stop. 
		3: A child laughed in the park. 
		4: A bird fluttered past the window. 
		5: The waves crashed against the shore. 
		6: A train whistled as it approached. 
		7: The fireworks lit up the sky. 
		8: A bicycle rode by. 
		9: A crowd cheered at the concert. \\
		\cmidrule(){2-2}
		&\multirow{1}{*}{\rotatebox{90}{user}}&The event you experienced is of type: \{ds\_event\_type\}. 
		In a longer text you are describing several things that happened at that event. 
		Something happened at that event with the following object(s): \{ds\_event\_object\}. 
		In your response, only provide a very short sentence describing what happened to/with the object(s).  \\
		\cmidrule(r){1-2}\cmidrule(l){3-3}
		
		\multirow{3}{*}{\rotatebox{90}{Prompt 2}}
		&\multirow{2}{*}{\rotatebox{90}{system}} 
		& 
		It is often clear from the text that describes an event which specific emotion it evokes in a person 
		that experienced it. However, additional information about the situation can change our understanding 
		of how a person might interpret the event. You are an expert at creating a scenario that explains 
		why a specific event may cause a possibly unusual emotion in you. In addition, you can concisely make 
		this scenario apparent for the reader by formulating a description of 4 events that took place 
		immediately before the event.\\
		\cmidrule(){2-2}
		&\multirow{1}{*}{\rotatebox{90}{user}}&You experienced something happening which is described by the following event description: 5. "\{event\}".\newline
		This event somehow made you clearly feel the emotion: "\{emotion\}".\newline
		Provide a text describing four events that took place immediately before event 5 by giving a 
		list of descriptions of these events (1.-4.). The events 1.-4. clearly influence your personal 
		emotional interpretation of the event that happened after (5.). 
		The emotion "\{emotion\}" is only triggered by what specifically happened in event 5. 
		The events 1.-4. evoked other emotions, such as: 
		\{", ".join(random.sample([emo for emo in EMOTION\_SET if emo != emotion], len(EMOTION\_SET)-1))\}.
		In your response, for each of the 4 event descriptions: Only give a summary text consisting 
		of the main clause in a very short sentence. Each description should only describe a singular 
		event. Indicate each event description in a separate line. \\
		\cmidrule(r){1-2}\cmidrule(l){3-3}
		
		\multirow{3}{*}{\rotatebox{90}{Prompt 2.1}}
		&\multirow{2}{*}{\rotatebox{90}{system}} 
		& 	It is often clear from the text that describes an event which specific emotion it evokes in a person 
		that experienced it. However, additional information about the situation can change our understanding 
		of how a person might interpret the event. You are an expert at creating a scenario that explains 
		why a specific event may cause a possibly unusual emotion in you. In addition, you can concisely make 
		this scenario apparent for the reader by formulating a description of 4 events that took place 
		immediately before the event. \\ 
		\cmidrule(){2-2}
		&\multirow{1}{*}{\rotatebox{90}{user}}&You experienced something happening which is described by the following event description: 5. "{event}".
		This event somehow made you clearly feel the emotion: "{emotion}".
		First, give a brief explanation of a scenario in which it can be deduced from the description of 
		event 5. that you felt {emotion}.
		Second, phrase this explanation as events that took place immediately before event 5 by giving a 
		list of descriptions of these events (1.-4.). The events 1.-4. clearly influence your personal 
		emotional interpretation of the event that happened after (5.). 
		The emotion "{emotion}" is only triggered by what specifically happened in event 5. 
		The events 1.-4. evoked other emotions, such as: 
		\{", ".join(random.sample([emo for emo in EMOTION\_SET if emo $\neq$ emotion], len(EMOTION\_SET)-1))\}. \\
		\cmidrule(r){1-2}\cmidrule(l){3-3}	
		
		\multirow{3}{*}{\rotatebox{90}{Pr. 2.2}}
		&\multirow{1}{*}{\rotatebox{90}{user}}&
		Extract the sequence of 4 descriptions of events that happened from the following text:
		\#\#\# \{explanation\} \#\#\#
		The event 5: "\{event\}" happened after the 4 events.
		In your response, for each of the 4 event descriptions: Only give a summary text consisting of the main clause in a very short sentence. Each description should only describe a singular event. Indicate each event description in a separate line.\\
		\cmidrule(r){1-2}\cmidrule(l){3-3}	
		
		\multirow{3}{*}{\rotatebox{90}{Prompt 2.3}}
		&\multirow{2}{*}{\rotatebox{90}{system}} 
		& 	You are an expert at adapting a narrative to convey specific emotional interpretations. You will receive a text that outlines a sequence of events as experienced by an individual. Additionally, there will be an explanation of how a particular emotion is triggered in this individual based on the final event. \\
		\cmidrule(){2-2}
		&\multirow{1}{*}{\rotatebox{90}{user}}&Explanation: \{story\_plan\}\newline
		Event sequence: \{chain\}\newline
		First, provide a brief evaluation on how the first four events (1.-4.) of the sequence could be adjusted 
		to form a coherent narrative which better aligns with the conclusion given in the explanation. 
		The text of the last event (5.) should remain as is.\newline
		Second, provide a revised event sequence that incorporates these adjustments while keeping the sentence 
		length for each event description similar. Each event description should consist only of a main clause in 
		a very short sentence. Do not explicitly mention the emotions felt. \\
		
		\bottomrule
	\end{tabularx}
	\caption{List of prompts used in our methods for event chain generation. The first column shows the numbers of the specific prompts as introduced in \Cref{fig:overview}. `MT' refers to the prompt message type as specified in the input for the instruction-tuned LLM.}\label{tab:ftprompts}
\end{table*}

\section{Descriptive Dataset Statistics}\label{app:stats}
To assess lexical diversity, we analyze the unigrams used in the backstories as well as Jaccard-based diversity scores.
Additionally, we analyze the coherence of event chains.

We perform tokenization for each event description by removing punctuation symbols, converting all text to lowercase, and splitting the resulting string on whitespace characters.
Event descriptions of the final events $s_5$ in our dataset consist of 15.4 tokens on average.
The length of the event descriptions in the backstories are shown in \Cref{tab:tokstats}.

\begin{table}\centering\small
	\begin{tabular}{l rrrrr}
		\toprule
		Method& $s_1$&$s_2$&$s_3$&$s_4$&$\Sigma$\\
		\cmidrule(r){1-1}\cmidrule(rl){2-2}\cmidrule(rl){3-3}\cmidrule(rl){4-4}\cmidrule(rl){5-5}\cmidrule(l){6-6}
		Baseline& 16.4& 16.5& 17.6& 19.4& 69.8\\
		PC& 12.2& 12.1& 12.2& 12.7& 49.2\\
		PCR&18.3& 19.0& 20.2& 21.6& 79.2\\
		\bottomrule
	\end{tabular}
	\caption{Text length in number of tokens of event descriptions in backstories generated by different methods comprising four sentences each.}
	\label{tab:tokstats}
\end{table}

\subsection{Analysis of Lexical Diversity} \label{sec:mld}
For each of the three data generation methods, we identify the most frequent unigrams (nouns) overall and among emotion-specific subsets.

Additionally, we compute a comparative lexical diversity measure using Jaccard-based diversity scores, calculated through pairwise comparisons of the word types utilized in the backstories generated by the different methods.
We define this diversity score between two instances \( b_i \) and \( b_j \) as \( D(b_i, b_j) = 1 - J(b_i, b_j) \).
The Jaccard coefficient \citep{jaccard} for these instances is defined as
\[
J(b_i, b_j) = \frac{|T_i \cap T_j|}{|T_i \cup T_j|}
\]
where \( T_i \) and \( T_j \) represent the sets of unique word types in instances \( b_i \) and \( b_j \), respectively.
With this we define the diversity in a set of instances, e.g.\ a subset of backstories \( B = \{b_1, \dots, b_n\}\), as
\[
D(B)  = 1-  \frac{\sum_{i=1}^{n} \sum_{j=1}^{n}J(b_i, b_j)}{n^2}.
\]
A high diversity score indicates that the backstories are highly different from each other, reflecting greater lexical diversity, while a low score suggests that the backstories share many common word types and are thus more similar.

\subsection{Coherence Scoring Algorithm}\label{sec:csa}
To assess the coherence of event chains, we adopt a zero-shot shuffle test methodology as suggested by \citet{laban-etal-2021-transformer}.
In the shuffle test, the texts are divided into smaller units, such as sentences, which are subsequently shuffled to create permutations.
A large language model is employed to evaluate the coherence of the original and each permutation by extracting the likelihood predictions for the token sequences.
Lower likelihood scores are interpreted to be indicative of permutations that exhibit reduced coherence.
The framework proposed by \citet{laban-etal-2021-transformer} highlights that employing a zero-shot setting, wherein the language model is utilized without prior fine-tuning specific to the shuffle test, facilitates a more accurate assessment of coherence.

We implement this methodology to compute coherence scores for each event chain \( c = (s_1, \dots, s_l) \) within our dataset, where \( l = 5 \).
The adapted algorithm is detailed as follows.

\paragraph{1. Shuffling.}
We include shuffled variants for each event chain  \( c \)  by computing the set of its permutations \( \mathfrak{S}_c = \{ \bar{c}_1, \dots , \bar{c}_k \}  \), where  \( \bar{c}_1 = c \) and \( k = l! \) (in our dataset,  \( k = 5! = 120 \)).
To reduce the computational cost, we randomly sample \( \frac{k}{4} = 30 \) permutations from this set, resulting in \( \mathfrak{S}'_c \), while ensuring that \( c\) is included in this sample.

\paragraph{2. Language Modeling.}
We encode the original event chain and the sampled shuffled variants using a language model \( M \)\footnote{We utilize the Llama-3.1-8B-Instruct model from the Meta-Llama collection, accessible via HuggingFace: \url{https://huggingface.co/meta-llama/Llama-3.1-8B-Instruct}.}.
\( M \) implements a function \( g \) which tokenizes\footnote{Before tokenization, the texts of the events in each event sequence are combined into a single string using a whitespace character as separator.} each \( \bar{c} \in \mathfrak{S}'_c \) into a sequence of tokens \( g (\bar{c}) = (t_1, \dots, t_n) \). 
For each tokenized sequence \( g (\bar{c}) \), \( M \) computes a sequence of prediction score vectors as output of its language modeling head, i.e.\ the logits of \( M \), as 
\( f(g (\bar{c})) = f((t_1, \dots, t_n)) = (l_1, \dots, l_n) \).
We transform these scores into values which we can interpret as log probabilities of transitions by applying a softmax followed by a logarithm for each token position as
\[
\begin{split}
	h_i &= \log(\text{Softmax}(l_i)) = [ \log(\frac{e^{l_{ij}}}{\sum_{q=1}^v e^{l_{iq}}}) ]
\end{split}
\]
for \( i = 1, \dots, n \) and  \( j = 1, \dots, v \), where \( v \) is the size of the vocabulary of  \( M \).
The log-likelihood of each sequence is computed based on the transition probabilities as
\[
\begin{split}
	\log{P(\bar{c})} &= \sum_{i=1}^{n} \log{P}(t_i | t_0 \dots t_{i-1}) = \sum_{i=1}^{n} h_{i, w(t_{i+1})}
\end{split}
\]
where \( w(t_{i+1}) \) is the index of \( t_{i+1} \in g (\bar{c}) \) in the vocabulary of \( M \) and \( h_{i, w} \) denotes the value in position \( w \) of the vector \( h_i \).

For the original event chain and each of its shuffled permutations \( \bar{c} \in \mathfrak{S}'_c \), we calculate their log-likelihood scores \( \log P( \bar{c}) \).

\paragraph{3. Coherence Score Calculation.}
We continue by ranking the original chain's likelihood against those of all its shuffled permutations.
The coherence score for the original chain is computed as 
\[
H(c) = 1 - \frac{r(c)}{| \mathfrak{S}'_c |}
\]
where \( r(c) \) is the rank of the original log-probability \( \log P(c) \) among the permutation scores, calculated as
\[
r(c) = \left| \{ \bar{c} \in \mathfrak{S}'_c \mid \log P(\bar{c}) \geq \log P(c) \} \right|.
\]

A higher score  \( H(c) \), approaching 1, indicates a high coherence of the event chain, while a score closer to 0 reflects low coherence.

\subsection{Results}\label{sec:res}
This section presents additional statistical analyses of our dataset, which consists of 13,000 event chains for each generation method.
Each event chain comprises five short texts, each a single sentence representing event descriptions from the perspective of an event experiencer, organized in chronological order.
The dataset is structured around 1,000 pivotal events, each enriched by four preceding events, collectively referred to as backstories, as detailed in \Cref{sec:gens}. 
Each pivotal event is paired with a unique backstory corresponding to one emotion, as outlined in \Cref{sec:def}, resulting in a comprehensive dataset that facilitates emotion analysis across various contexts. 

\paragraph{Unigrams.}
The top 20 unigrams in the backstories generated by the different methods are shown in \Cref{tab:unigrams}.

Additionally, \Cref{tab:unigramsu} shows unique items within the top 100 unigrams pertinent to each emotion category for each method.
This table reveals further distinctive characteristics tied to specific emotional narratives.
For instance, in the anger category, ``changes'' may suggest conflicts or disruptions, while a word such as ``wedding'' in the sadness category evokes nostalgic themes and significant life events.
Additionally, the term ``proposal'' in the pride category suggests achievement and recognition, contrasting sharply with ``trash'' and ``scandal'' in the disgust category, which imply negative experiences.
While some terms may appear less characteristic of the emotions, several others resonate clearly with the themes we expect to find in stories related to the corresponding emotions.

\begin{table*}[t!]
	\centering\scriptsize\sffamily
	\renewcommand{\arraystretch}{1.1} \setlength{\tabcolsep}{3pt}
	\begin{tabularx}{\textwidth}{p{.8cm}XXX}
		\toprule
		Subset& Baseline&PC&PCR\\ 
		\cmidrule(r){1-1}\cmidrule(lr){2-2}\cmidrule(lr){3-3}\cmidrule(l){4-4}
		anger&event, hours, organizer, friend, day, team, conference, months, party, morning, call, project, family, group, staff, presentation, organizers, sister, hour, front&event, team, organizers, organizer, staff, group, friend, family, morning, email, call, hours, venue, day, instructor, meeting, conference, project, member, party&event, organizers, team, hours, family, staff, organizer, group, friend, concerns, time, venue, day, project, email, conference, party, teams, room, stage\\
		boredom&friend, hours, event, hour, day, presentation, minutes, call, conference, project, argument, organizer, morning, party, group, time, team, coffee, family, front&friend, event, team, project, group, argument, organizer, call, coffee, family, instructor, conference, presentation, hours, venue, morning, stage, organizers, \textbf{train}, staff&event, time, friend, team, hours, stage, presentation, family, group, day, \textbf{crowd}, organizers, project, teams, conference, room, organizer, argument, venue, morning\\
		disgust&event, friend, organizer, hours, group, \textbf{food}, conference, \textbf{company}, day, morning, hour, staff, party, family, organizers, project, team, room, speech, front&event, organizer, friend, staff, organizers, group, team, \textbf{food}, family, conversation, room, member, venue, conference, stage, \textbf{companys}, call, meeting, morning, instructor&event, organizers, \textbf{food}, staff, organizer, friend, group, family, team, concerns, room, \textbf{hands}, conference, equipment, \textbf{trash}, member, \textbf{companys}, stage, venue, \textbf{company}\\
		fear&event, call, friend, day, \textbf{warning}, family, organizer, conference, morning, hours, group, email, \textbf{fire}, project, \textbf{phone}, team, \textbf{area}, \textbf{message}, \textbf{performance}, presentation&call, friend, group, event, team, family, email, conversation, project, \textbf{message}, colleague, stage, room, meeting, organizer, argument, venue, presentation, morning, organizers&family, event, team, friend, group, room, feeling, project, \textbf{eyes}, stage, call, email, \textbf{message}, \textbf{mind}, time, \textbf{performance}, presentation, teams, \textbf{safety}, venue\\
		guilt&friend, event, day, sister, organizer, family, team, morning, party, project, call, group, conference, time, \textbf{colleague}, months, hours, work, night, presentation&friend, event, team, call, colleague, family, project, friends, morning, email, organizer, group, conversation, party, argument, meeting, organizers, day, concerns, room&friend, event, team, friends, family, project, colleague, time, concerns, group, call, party, organizers, day, feeling, organizer, morning, \textbf{task}, attendees, work\\
		joy&friend, hours, event, day, family, call, morning, sister, argument, months, project, team, conference, time, email, hour, party, organizer, presentation, coffee&team, friend, family, event, hours, group, venue, email, morning, call, project, argument, conversation, stage, day, sound, \textbf{teams}, party, room, time&team, event, family, friend, time, hours, venue, room, day, teams, stage, party, project, group, sound, morning, friends, email, call, issues\\
		pride&event, team, hours, morning, project, months, family, night, day, time, organizer, friend, group, conference, teams, party, presentation, sister, work, organizers&team, event, hours, call, group, email, friend, concerns, morning, venue, \textbf{system}, family, meeting, \textbf{night}, equipment, member, organizers, argument, \textbf{design}, conference&team, event, hours, concerns, time, teams, \textbf{system}, group, family, \textbf{design}, issue, \textbf{plan}, work, project, friend, venue, equipment, attendees, meeting, party\\
		relief&hours, event, day, call, organizer, time, team, morning, argument, hour, friend, minutes, conference, presentation, night, family, party, group, sister, front&team, call, event, friend, group, hours, family, room, stage, member, argument, time, morning, staff, sound, equipment, project, instructor, venue, meeting&team, time, event, hours, room, family, stage, venue, equipment, group, call, sound, friend, attendees, concerns, schedule, teams, argument, project, \textbf{solution}\\
		sadness&friend, call, family, sister, event, \textbf{grandmother}, day, morning, months, \textbf{years}, hours, party, conference, team, organizer, project, argument, \textbf{concert}, time, \textbf{grandmothers}&friend, family, call, event, \textbf{grandmothers}, \textbf{grandmother}, venue, morning, team, group, friends, email, conversation, \textbf{photo}, \textbf{grandfathers}, project, party, day, \textbf{sister}, organizers&friend, \textbf{grandmother}, family, \textbf{grandmothers}, event, \textbf{memories}, \textbf{grandfather}, friends, time, \textbf{grandfathers}, team, call, venue, hours, \textbf{photo}, project, group, morning, day, stage\\
		shame&event, friend, organizer, family, day, group, team, presentation, \textbf{friends}, morning, front, party, hours, months, \textbf{skills}, conference, project, organizers, speech, sister&friend, friends, team, event, group, family, call, colleague, organizer, project, party, email, morning, conference, presentation, day, \textbf{skills}, venue, organizers, argument&friend, event, friends, team, group, family, time, party, project, colleague, \textbf{skills}, morning, day, organizer, conference, organizers, everything, attendees, feeling, concerns\\
		surprise&event, hours, organizer, friend, day, morning, conference, staff, party, minutes, hour, team, room, family, stage, argument, time, \textbf{issues}, presentation, organizers&event, team, friend, organizer, organizers, staff, group, venue, argument, instructor, room, stage, family, conference, email, member, conversation, call, morning, \textbf{attendees}&event, team, staff, organizers, organizer, room, friend, venue, stage, time, group, attendees, family, hours, conference, teams, equipment, concerns, instructor, sound\\
		trust&friend, event, organizer, hours, staff, group, team, argument, family, sister, party, conference, day, call, \textbf{instructor}, presentation, teams, \textbf{speaker}, morning, stage&team, friend, event, staff, group, organizers, argument, instructor, venue, meeting, family, member, concerns, stage, \textbf{leader}, organizer, project, hours, \textbf{line}, conversation&team, staff, event, friend, organizers, concerns, group, family, stage, time, venue, member, instructor, issues, issue, meeting, teams, project, room, sound\\
		no-emotion&event, friend, hours, day, call, hour, argument, family, organizer, conference, party, project, team, time, morning, work, minutes, coffee, \textbf{boss}, group&team, call, friend, event, argument, project, group, morning, email, \textbf{issue}, family, organizer, stage, colleague, time, equipment, coffee, member, venue, presentation&team, event, friend, argument, hours, time, issue, call, schedule, project, group, family, room, equipment, venue, stage, morning, everything, presentation, sound\\
		\cmidrule(r){1-1}\cmidrule(lr){2-2}\cmidrule(lr){3-3}\cmidrule(l){4-4}Overall&event, friend, hours, day, organizer, family, morning, team, call, conference, party, group, project, sister, hour, presentation, months, argument, time, staff&team, friend, event, group, call, family, organizer, project, venue, email, morning, organizers, staff, argument, hours, instructor, conversation, stage, conference, meeting&event, team, friend, family, time, group, hours, organizers, room, project, staff, stage, venue, concerns, organizer, day, teams, friends, conference, call\\
		\bottomrule
	\end{tabularx}
	\caption{Top 20 unigrams (nouns only) in backstories generated by different methods. Unigrams marked in bold are unique for the set of chains of the respective emotion.}\label{tab:unigrams}
\end{table*}

\begin{table*}\centering\scriptsize
	\renewcommand{\arraystretch}{1.2}\sffamily
	\begin{tabularx}{\textwidth}{p{1cm}XXX}
		\toprule
		Subset& Baseline&PC&PCR\\ 
		\cmidrule(r){1-1}\cmidrule(lr){2-2}\cmidrule(lr){3-3}\cmidrule(l){4-4}
		anger&---&changes, weekend, booth&changes\\
		boredom&tray, guest, spot&performer, alarm, keynote, drinks, course, lecture, blew, opening, drink, minute&act, routine, week, break, keynote, alarm\\
		disgust&trash, sandwich, impact, festivals, sponsor, field&hands, scandal, article, bag, practices, history, spill, document&hands, trash, smell, table, history, scandal, floor, tone, health, waste, practices, bag, signs, person, dirty, impact, sustainability\\
		fear&message, security, number, venues, history, department, workshop, accident&security, classmate, sign, child, office&warning, something, security, job, expression\\
		guilt&struggles, gift, responsibility, help, sisters, decision&task, sibling, care, budget&decision, responsibility, glimpse, disappointment, media, tasks, sibling, didnt\\
		joy&---&pride&anticipation\\
		pride&design, ability, proposal, setup, volunteer&ability, kids, idea, management, lastminute, encouragement&confidence, ability, proposal, action, determination, sponsor, abilities, program, management, guidance\\
		relief&deadline, planner, backup&supplier, solution, delivery, traffic, situation, rehearsal, couldnt&supplier, traffic, frustration\\
		sadness&grandmothers, grandfather, health, grandfathers, photo, father&grandmothers, photo, grandfathers, grandfather, mother, heartfelt, photos, belongings, invitation, house, school, injury, note, name, box, band&grandmothers, grandfather, grandfathers, photo, stories, life, photos, memory, years, mother, celebration, share, contrast, note, father, wedding, countless, heartfelt, days\\
		shame&coach, art, classmates, colleagues, name, aunt&skills, everyone, attention, importance, progress, concern&expertise, classmates, someone, control\\
		surprise&weather, forecast, crew&presenter, performers&arrival, crew, performers, doubts\\
		trust&smile, stranger, kind, apology, passenger, mistakes&mistakes, process, entrance, vendor&entrance, questions, ticket\\
		no-emotion&times, plans&home, nod, supervisor, stand, speakers&workshop, home, text, weight, emergency\\
		\bottomrule
	\end{tabularx}
	\caption{Unique words for each emotion category in the top 100 unigrams (nouns only) in backstories generated by different methods.}\label{tab:unigramsu}
\end{table*}

\paragraph{Lexical Diversity based on Jaccard Coefficient.}

\begin{table}\centering\scriptsize
	\begin{tabular}{l rrr}
		\toprule
		Subset& Baseline&PC&PCR\\
		\cmidrule(r){1-1}\cmidrule(rl){2-2}\cmidrule(rl){3-3}\cmidrule(l){4-4}
		anger& .87& .90& .89\\
		boredom& .87& .90& .90\\
		disgust& .87& .90& .90\\
		fear& .87& .89& .89\\
		guilt& .87& .89& .89\\
		joy& .86& .90& .90\\
		pride& .87& .90& .90\\
		relief& .87& .91& .90\\
		sadness& .86& .89& .89\\
		shame& .87& .89& .89\\
		surprise& .87& .91& .91\\
		trust& .88& .90& .90\\
		no-emotion& .86& .89& .90\\
		\cmidrule(r){1-1}\cmidrule(rl){2-2}\cmidrule(rl){3-3}\cmidrule(l){4-4}
		Overall& .87& .90& .90\\
		\bottomrule
	\end{tabular}
	\caption{Average pairwise Jaccard-based diversity scores of backstories by different methods.}
	\label{tab:jdivstats}
\end{table}

We further analyze lexical diversity using average pairwise Jaccard-based diversity scores, summarized in \Cref{tab:jdivstats}.
For clarity, the Jaccard coefficient allows us to yield diversity scores that reflect the degree of dissimilarity between backstories.
A high (or low) diversity score indicates that instances are very different from (or similar to) each other, reflecting correspondingly high (or low) diversity.
Overall, this analysis indicates a consistent level of lexical diversity throughout the dataset.
The overall lack of biases in lexical diversity across different emotion categories further underscores the homogeneity of the data, implying that the narratives were constructed as diverse as intended without influence from the emotion categorization.

\section{Examples of Generated Event Chains}\label{app:examples}

\Cref{tab:examplesm} shows narratives generated by our three methods for four emotion categories.
On closer inspection, we can see that, compared to the other approaches, the Baseline tends to link the backstory very strongly to the last event and frequently reuse content mentioned in it.
In contrast, the narratives of the other two methods seem to be more complex and go further into the background.
The PCR method in particular conveys the narrative more clearly by emphasizing the backstory more elaborately in the sequence of events.
Overall, this illustrates how the content derived from the approach PC is adopted and enriched in various spots.

\begin{table*}[t!]\centering
	\centering\sffamily\scriptsize
	\renewcommand{\arraystretch}{.75}\setlength{\tabcolsep}{2.5pt}
	\begin{tabularx}{\linewidth}{@{\hspace{1.mm}}p{4mm}X@{\hspace{3.mm}}X@{\hspace{3.mm}}X@{\hspace{1.mm}}}
		\toprule
		Em.&\multicolumn{1}{c}{Baseline}&\multicolumn{1}{c}{PC}&\multicolumn{1}{c}{PCR}\\
		\cmidrule(r){1-1}\cmidrule(rl){2-2}\cmidrule(rl){3-3}\cmidrule(l){4-4}
		\multirow{11}{*}{\rotatebox{90}{Guilt}}
		& 
		I had been tasked with testing the loudspeaker system before the big event.
		My supervisor warned me that a malfunction would be disastrous for the company's reputation.
		I skipped the recommended final check to grab a quick lunch before the event started.
		The event host began the ceremony, and the loudspeaker was working perfectly, filling me with temporary relief.
		\textit{The loudspeaker suddenly malfunctioned and went silent.}
		& 
		I was tasked with setting up the sound system for a charity event.
		I noticed the previous technician had left a mess with wires and cables scattered all over the floor.
		I decided to skip some of the safety checks to meet the deadline.
		I received a call from my friend, reminding me how much the event meant to her.
		\textit{The loudspeaker suddenly malfunctioned and went silent.}
		& 
		A close friend entrusted me with setting up the sound system for a charity event.
		I faced a daunting task, as the previous technician had left the equipment in disarray.
		With time running out, I took a shortcut to meet the deadline, skipping some crucial safety checks.
		Just before the event started, my friend reminded me of its significance and the many people counting on its success.
		\textit{The loudspeaker suddenly malfunctioned and went silent.}\\
		
		\cmidrule(r){1-1}\cmidrule(rl){2-2}\cmidrule(rl){3-3}\cmidrule(l){4-4}
		\multirow{12}{*}{\rotatebox{90}{Relief}}
		& 
		The loudspeaker had been blasting earsplitting static for 10 minutes.
		I had just received an urgent phone call from the event organizer, warning me of a severe thunderstorm approaching.
		The previous speaker's presentation was a disaster, with him stumbling over his words and spilling water on his notes.
		I had spent the last hour setting up the outdoor stage, and a strong gust of wind had just blown over a music stand.
		\textit{The loudspeaker suddenly malfunctioned and went silent.}
		& 
		I was tasked with giving a presentation to a large crowd.
		The sound system malfunctioned, amplifying my voice to an ear-piercing level.
		The sound technician ignored the problem and chatted with someone.
		The audience covered their ears and looked at me with discomfort.
		\textit{The loudspeaker suddenly malfunctioned and went silent.}
		& 
		I stood before a sea of expectant faces, the weight of their attention crushing me.
		The sound system suddenly shrieked to life, my amplified voice piercing the air like a knife.
		The sound technician fumbled with the controls, oblivious to the cacophony they were creating.
		The audience winced in unison, their pained expressions mirroring the turmoil brewing inside me.
		\textit{The loudspeaker suddenly malfunctioned and went silent.}\\
		
		\cmidrule(r){1-1}\cmidrule(rl){2-2}\cmidrule(rl){3-3}\cmidrule(l){4-4}
		\multirow{16}{*}{\rotatebox{90}{Fear}}
		& 
		I had just given a speech to a large crowd, criticizing the company's management.
		The CEO, visibly angry, approached the microphone to respond to my accusations.
		The loudspeaker suddenly screeched, causing the audience to cover their ears in discomfort.
		The CEO's face turned red as he began to speak, but his voice was barely audible due to the screeching.
		\textit{The loudspeaker suddenly malfunctioned and went silent.}
		& 
		I arrived at a remote wilderness survival training camp.
		I learned the camp was located near a toxic waste site.
		A fellow trainee struggled with the physical demands and dropped out.
		I completed a challenging obstacle course.
		\textit{The loudspeaker suddenly malfunctioned and went silent.}
		& 
		I arrived at a remote wilderness survival training camp, where the instructors emphasized the importance of following loudspeaker instructions for safety.
		The instructors warned us about the toxic waste site nearby and explained that the loudspeaker would alert us to any changes in air quality.
		During the first exercise, I struggled to navigate the challenging terrain, but the loudspeaker provided crucial guidance, helping me stay on track.
		I completed a difficult obstacle course, relying heavily on the loudspeaker's instructions to avoid hazards and find the safest route.
		\textit{The loudspeaker suddenly malfunctioned and went silent.}\\
		\cmidrule(r){1-1}\cmidrule(rl){2-2}\cmidrule(rl){3-3}\cmidrule(l){4-4}
		\multirow{14}{*}{\rotatebox{90}{Pride}}
		& 
		I had just finished a long and difficult speech in front of a large audience.
		The loudspeaker had been malfunctioning throughout my presentation, causing me to struggle to be heard.
		The audience was getting restless and some people started to leave due to the poor sound quality.
		I had just made a crucial point that seemed to resonate with the remaining audience members.
		\textit{The loudspeaker suddenly malfunctioned and went silent.}
		& 
		I set up the sound system for the big conference.
		The event organizer informed me that the conference was running 30 minutes behind schedule.
		I listened to the same annoying music loop playing over and over again.
		The keynote speaker began to talk with the sound system working flawlessly.
		\textit{The loudspeaker suddenly malfunctioned and went silent.}
		& 
		I spent the entire morning upgrading the sound system with a new backup system to prevent technical issues.
		The event organizer informed me that the conference was running 30 minutes behind schedule, giving me extra time to test the new backup system.
		I used the extra time to run a series of tests on the sound system, trying to simulate potential failures.
		The keynote speaker began to talk, and the sound system was working flawlessly, but I was still waiting for a real test of the new backup system.
		\textit{The loudspeaker suddenly malfunctioned and went silent.}\\
		
		\bottomrule
	\end{tabularx}
	\caption{Examples of event chains generated by different methods.}\label{tab:examplesm}
\end{table*}

\section{Human Annotation Setup and Analysis}\label{app:anno}

We performed a human annotation of events and event chains with regard to emotion analysis and text quality.
We use the platform Prolific\footnote{\url{https://www.prolific.com/}} to find study participants.
These are screened to be in the UK, have English as their first and native language, and have an approval rate of at least 95\%.

\subsection{Annotation of Events}

We sample a set of 10 events (one from each event type category, cf. \Cref{app:eventtypes}) and collect 39 annotations for this sample.\footnote{We collect 39 annotations for each event since we intend to compare these to the annotations of 13 corresponding event chains (with 3 annotations each).}
Additionally, we sample 90 more events (nine from each event type category, cf. \Cref{app:eventtypes}) and collect 3 annotations each, so we can evaluate the quality of the generated texts on a sample of 100 events (10\% of our dataset).

\subsection{Annotation of Event Chains}

We collect 780 annotations of event chains, which corresponds to three annotations per 13 backstories, generated by each of the two methods, for each of the 10 sampled events.
\Cref{tab:pe_annoe_m_bc} shows the annotated labels in comparison to the labels in the prompts with the baseline method.
This annotation results in the evaluation scores as shown in \Cref{tab:sc_pe_annoe_m_bc}.
The comparison of the annotated labels to the labels in the prompts with the PCR method results in the evaluation scores as shown in \Cref{tab:sc_pe_annoe_m_igc}.
We also ask annotators to evaluate the quality of the narratives based on a five-point Likert scale regarding the contextual influence of the backstory, realism of the story and whether it is written by a human or by an AI.
Results for this annotation (see \Cref{tab:annos}) do not show clear differences between the two tested methods.

\begin{table}
	\centering\small \setlength{\tabcolsep}{1.5pt}
	\begin{tabular}{l@{\hspace{.75\tabcolsep}}rrrrrrrrrrrrr}
		\toprule
		\diagbox[innerleftsep=0cm,width=1.25cm,height=1.4cm]{$e(b{\mkern1.5mu{\cdot}\mkern1.5mu}s_5)$}{$e_A$}& \rotatebox[origin=c]{90}{Anger}&\rotatebox[origin=c]{90}{Boredom}&\rotatebox[origin=c]{90}{Disgust}&\rotatebox[origin=c]{90}{Fear}&\rotatebox[origin=c]{90}{Guilt}&\rotatebox[origin=c]{90}{Joy}&\rotatebox[origin=c]{90}{Pride}&\rotatebox[origin=c]{90}{Relief}&\rotatebox[origin=c]{90}{Sadness}&\rotatebox[origin=c]{90}{Shame}&\rotatebox[origin=c]{90}{Surprise}&\rotatebox[origin=c]{90}{Trust}&\rotatebox[origin=c]{90}{No-Emot.}\\
		\cmidrule(r{.3em}){1-1}\cmidrule(r{.3em}l{.3em}){2-2}\cmidrule(r{.3em}l{.3em}){3-3}\cmidrule(r{.3em}l{.3em}){4-4}\cmidrule(r{.3em}l{.3em}){5-5}\cmidrule(r{.3em}l{.3em}){6-6}\cmidrule(r{.3em}l{.3em}){7-7}\cmidrule(r{.3em}l{.3em}){8-8}\cmidrule(r{.3em}l{.3em}){9-9}\cmidrule(r{.3em}l{.3em}){10-10}\cmidrule(r{.3em}l{.3em}){11-11}\cmidrule(r{.3em}l{.3em}){12-12}\cmidrule(r{.3em}l{.3em}){13-13}\cmidrule(l{.3em}){14-14}
		
		Anger& \cellcolor{blue!33}10& \cellcolor{blue!0}0& \cellcolor{blue!13}4& \cellcolor{blue!0}0& \cellcolor{blue!3}1& \cellcolor{blue!0}0& \cellcolor{blue!0}0& \cellcolor{blue!10}3& \cellcolor{blue!6}2& \cellcolor{blue!3}1& \cellcolor{blue!3}1& \cellcolor{blue!3}1& \cellcolor{blue!23}7\\
		Boredom& \cellcolor{blue!23}7& \cellcolor{blue!10}3& \cellcolor{blue!3}1& \cellcolor{blue!6}2& \cellcolor{blue!0}0& \cellcolor{blue!13}4& \cellcolor{blue!3}1& \cellcolor{blue!3}1& \cellcolor{blue!3}1& \cellcolor{blue!3}1& \cellcolor{blue!10}3& \cellcolor{blue!6}2& \cellcolor{blue!13}4\\
		Disgust& \cellcolor{blue!30}9& \cellcolor{blue!0}0& \cellcolor{blue!6}2& \cellcolor{blue!0}0& \cellcolor{blue!6}2& \cellcolor{blue!3}1& \cellcolor{blue!0}0& \cellcolor{blue!3}1& \cellcolor{blue!10}3& \cellcolor{blue!0}0& \cellcolor{blue!16}5& \cellcolor{blue!0}0& \cellcolor{blue!23}7\\
		Fear& \cellcolor{blue!13}4& \cellcolor{blue!0}0& \cellcolor{blue!3}1& \cellcolor{blue!30}9& \cellcolor{blue!6}2& \cellcolor{blue!3}1& \cellcolor{blue!3}1& \cellcolor{blue!6}2& \cellcolor{blue!13}4& \cellcolor{blue!0}0& \cellcolor{blue!3}1& \cellcolor{blue!0}0& \cellcolor{blue!16}5\\
		Guilt& \cellcolor{blue!13}4& \cellcolor{blue!3}1& \cellcolor{blue!0}0& \cellcolor{blue!3}1& \cellcolor{blue!13}4& \cellcolor{blue!3}1& \cellcolor{blue!0}0& \cellcolor{blue!10}3& \cellcolor{blue!20}6& \cellcolor{blue!3}1& \cellcolor{blue!3}1& \cellcolor{blue!3}1& \cellcolor{blue!23}7\\
		Joy& \cellcolor{blue!6}2& \cellcolor{blue!6}2& \cellcolor{blue!3}1& \cellcolor{blue!10}3& \cellcolor{blue!3}1& \cellcolor{blue!13}4& \cellcolor{blue!3}1& \cellcolor{blue!13}4& \cellcolor{blue!23}7& \cellcolor{blue!3}1& \cellcolor{blue!6}2& \cellcolor{blue!0}0& \cellcolor{blue!6}2\\
		Pride& \cellcolor{blue!10}3& \cellcolor{blue!3}1& \cellcolor{blue!3}1& \cellcolor{blue!3}1& \cellcolor{blue!6}2& \cellcolor{blue!3}1& \cellcolor{blue!30}9& \cellcolor{blue!10}3& \cellcolor{blue!6}2& \cellcolor{blue!0}0& \cellcolor{blue!3}1& \cellcolor{blue!6}2& \cellcolor{blue!13}4\\
		Relief& \cellcolor{blue!20}6& \cellcolor{blue!6}2& \cellcolor{blue!0}0& \cellcolor{blue!10}3& \cellcolor{blue!0}0& \cellcolor{blue!3}1& \cellcolor{blue!0}0& \cellcolor{blue!13}4& \cellcolor{blue!20}6& \cellcolor{blue!6}2& \cellcolor{blue!13}4& \cellcolor{blue!0}0& \cellcolor{blue!6}2\\
		Sadness& \cellcolor{blue!10}3& \cellcolor{blue!6}2& \cellcolor{blue!3}1& \cellcolor{blue!6}2& \cellcolor{blue!3}1& \cellcolor{blue!3}1& \cellcolor{blue!3}1& \cellcolor{blue!3}1& \cellcolor{blue!23}7& \cellcolor{blue!0}0& \cellcolor{blue!10}3& \cellcolor{blue!10}3& \cellcolor{blue!16}5\\
		Shame& \cellcolor{blue!3}1& \cellcolor{blue!6}2& \cellcolor{blue!6}2& \cellcolor{blue!3}1& \cellcolor{blue!10}3& \cellcolor{blue!0}0& \cellcolor{blue!10}3& \cellcolor{blue!10}3& \cellcolor{blue!6}2& \cellcolor{blue!23}7& \cellcolor{blue!3}1& \cellcolor{blue!0}0& \cellcolor{blue!16}5\\
		Surprise& \cellcolor{blue!20}6& \cellcolor{blue!3}1& \cellcolor{blue!3}1& \cellcolor{blue!6}2& \cellcolor{blue!0}0& \cellcolor{blue!13}4& \cellcolor{blue!3}1& \cellcolor{blue!16}5& \cellcolor{blue!6}2& \cellcolor{blue!0}0& \cellcolor{blue!20}6& \cellcolor{blue!0}0& \cellcolor{blue!6}2\\
		Trust& \cellcolor{blue!13}4& \cellcolor{blue!3}1& \cellcolor{blue!3}1& \cellcolor{blue!13}4& \cellcolor{blue!3}1& \cellcolor{blue!3}1& \cellcolor{blue!3}1& \cellcolor{blue!10}3& \cellcolor{blue!10}3& \cellcolor{blue!0}0& \cellcolor{blue!13}4& \cellcolor{blue!6}2& \cellcolor{blue!16}5\\
		No-Emot.& \cellcolor{blue!13}4& \cellcolor{blue!3}1& \cellcolor{blue!0}0& \cellcolor{blue!6}2& \cellcolor{blue!3}1& \cellcolor{blue!0}0& \cellcolor{blue!0}0& \cellcolor{blue!13}4& \cellcolor{blue!30}9& \cellcolor{blue!3}1& \cellcolor{blue!6}2& \cellcolor{blue!0}0& \cellcolor{blue!20}6\\
		$\Sigma$&63&16&15&30&18&19&18&37&54&14&34&11&61\\
		\bottomrule
	\end{tabular}
	\caption{Counts of prompted emotion $e(b{\mkern1.5mu{\cdot}\mkern1.5mu}s_5)$ vs.\ annotated emotion $e_A$ for event chains (generation method: Baseline).}
	\label{tab:pe_annoe_m_bc}
\end{table}

\begin{table}
	\centering\scriptsize \setlength{\tabcolsep}{1.1pt}
	\begin{tabular}{l rrrrrrrrrrrrr r}
		\toprule
		&\rotatebox[origin=c]{90}{Anger}&\rotatebox[origin=c]{90}{Boredom}&\rotatebox[origin=c]{90}{Disgust}&\rotatebox[origin=c]{90}{Fear}&\rotatebox[origin=c]{90}{Guilt}&\rotatebox[origin=c]{90}{Joy}&\rotatebox[origin=c]{90}{Pride}&\rotatebox[origin=c]{90}{Relief}&\rotatebox[origin=c]{90}{Sadness}&\rotatebox[origin=c]{90}{Shame}&\rotatebox[origin=c]{90}{Surprise}&\rotatebox[origin=c]{90}{Trust}&\rotatebox[origin=c]{90}{No-Emot.}&Ma.-Avg.\\
		\cmidrule(r{.3em}){1-1}\cmidrule(r{.3em}l{.3em}){2-2}\cmidrule(r{.3em}l{.3em}){3-3}\cmidrule(r{.3em}l{.3em}){4-4}\cmidrule(r{.3em}l{.3em}){5-5}\cmidrule(r{.3em}l{.3em}){6-6}\cmidrule(r{.3em}l{.3em}){7-7}\cmidrule(r{.3em}l{.3em}){8-8}\cmidrule(r{.3em}l{.3em}){9-9}\cmidrule(r{.3em}l{.3em}){10-10}\cmidrule(r{.3em}l{.3em}){11-11}\cmidrule(r{.3em}l{.3em}){12-12}\cmidrule(r{.3em}l{.3em}){13-13}\cmidrule(r{.3em}l{.3em}){14-14}\cmidrule(l{.3em}){15-15}
		Precision& .33& .10& .07& .30& .13& .13& .30& .13& .23& .23& .20& .07& .20& .19\\
		Recall& .16& .19& .13& .30& .22& .21& .50& .11& .13& .50& .18& .18& .10& .22\\
		F1& .22& .13& .09& .30& .17& .16& .37& .12& .17& .32& .19& .10& .13& .19\\
		\bottomrule
	\end{tabular}
	\caption{Scores for prompted emotion vs.\ annotated emotion for event chain (generation method: Baseline). Ma.-Avg.: Macro-Average.}
	\label{tab:sc_pe_annoe_m_bc}
\end{table}

\begin{table}
	\centering\scriptsize \setlength{\tabcolsep}{1.1pt}
	\begin{tabular}{l rrrrrrrrrrrrr r}
		\toprule
		&\rotatebox[origin=c]{90}{Anger}&\rotatebox[origin=c]{90}{Boredom}&\rotatebox[origin=c]{90}{Disgust}&\rotatebox[origin=c]{90}{Fear}&\rotatebox[origin=c]{90}{Guilt}&\rotatebox[origin=c]{90}{Joy}&\rotatebox[origin=c]{90}{Pride}&\rotatebox[origin=c]{90}{Relief}&\rotatebox[origin=c]{90}{Sadness}&\rotatebox[origin=c]{90}{Shame}&\rotatebox[origin=c]{90}{Surprise}&\rotatebox[origin=c]{90}{Trust}&\rotatebox[origin=c]{90}{No-Emot.}&Ma.-Avg.\\
		\cmidrule(r{.3em}){1-1}\cmidrule(r{.3em}l{.3em}){2-2}\cmidrule(r{.3em}l{.3em}){3-3}\cmidrule(r{.3em}l{.3em}){4-4}\cmidrule(r{.3em}l{.3em}){5-5}\cmidrule(r{.3em}l{.3em}){6-6}\cmidrule(r{.3em}l{.3em}){7-7}\cmidrule(r{.3em}l{.3em}){8-8}\cmidrule(r{.3em}l{.3em}){9-9}\cmidrule(r{.3em}l{.3em}){10-10}\cmidrule(r{.3em}l{.3em}){11-11}\cmidrule(r{.3em}l{.3em}){12-12}\cmidrule(r{.3em}l{.3em}){13-13}\cmidrule(r{.3em}l{.3em}){14-14}\cmidrule(l{.3em}){15-15}
		Precision& .17& .00& .20& .40& .20& .07& .33& .20& .50& .37& .20& .07& .17& .22\\
		Recall& .12& .00& .55& .34& .50& .11& .27& .13& .24& .50& .20& .20& .08& .25\\
		F1& .14& .00& .29& .37& .29& .08& .30& .16& .33& .42& .20& .10& .11& .21\\
		\bottomrule
	\end{tabular}
	\caption{Scores for prompted emotion vs.\ annotated emotion for event chain (generation method: PCR). Ma.-Avg.: Macro-Average.}
	\label{tab:sc_pe_annoe_m_igc}
\end{table}

\begin{table}
	\centering\small \setlength{\tabcolsep}{4pt}
	\begin{tabular}{l rr r rr}
		\toprule
		&\multicolumn{2}{c}{Infl.}&&\multicolumn{2}{c}{Written By}\\
		\cmidrule(){2-3}\cmidrule(){5-6}
		&Yes&No&Real.&Human&AI\\
		\cmidrule(r){2-2}\cmidrule(rl){3-3}\cmidrule(rl){4-4}\cmidrule(rl){5-5}\cmidrule(l){6-6}
		Baseline&249&141& 3.56& 2.98& 3.23\\
		PCR&264&126& 3.52& 2.95& 3.21\\
		\bottomrule
	\end{tabular}
	\caption{Human evaluation of the quality of the generated narratives: Binary values for the contextual influence of the backstory (Infl.); averaged five-point Likert scale values regarding the plausibility of the story (Real.) and whether it is written by a human or by an AI.}
	\label{tab:annos}
\end{table}

\section{Correlation of Human and System Emotion Annotation} \label{app:corrclf}
The correlations between human annotations and system predictions are summarized in \Cref{tab:corrclf}.
These findings are derived from the evaluation of 100 event annotations and 260 chain annotations (all with 3 annotators each).
The correlations values for events exhibit greater fluctuations across categories, whereas those for chains are more consistent.
The findings reveal that, for individual events, there are notable differences in the understanding of categories such as disgust, guilt, and shame between human annotators and the system, as evidenced by the low correlation in these categories.
Conversely, for the chains, only the emotion category of boredom shows a disparity in recognition between human annotators and the system, indicated by similarly low correlation.
The highest correlation coefficients for events are observed in the categories of joy, surprise, and no-emotion, while for chains, highest correlations are noted for anger, fear, pride, and sadness.

\begin{table}
	\setlength{\tabcolsep}{3.75pt}
	\centering\scriptsize
	\begin{tabular}{p{2mm} l r@{\hspace{.5\tabcolsep}}l r@{\hspace{.5\tabcolsep}}l}
		\toprule
		&&\multicolumn{2}{l}{Events}&\multicolumn{2}{l}{Chains}\\
		\cmidrule(r){3-4} \cmidrule(l){5-6}
		\multirow{13}{*}{\rotatebox{90}{Emotion-Specific Annotation}}
		&Anger& .46&***& .44&***\\
		&Boredom& .29&**& .15&*\\
		&Disgust& .14&& .22&***\\
		&Fear& .39&***& .46&***\\
		&Guilt& .10&& .27&***\\
		&Joy& .60&***& .31&***\\
		&Pride& .41&***& .43&***\\
		&Relief& .43&***& .38&***\\
		&Sadness& .30&**& .49&***\\
		&Shame& .13&& .38&***\\
		&Surprise& .50&***& .34&***\\
		&Trust& .37&***& .29&***\\
		&No-Emotion& .55&***& .22&***\\
		\cmidrule(r){1-2} \cmidrule(r){3-4} \cmidrule(l){5-6}
		\multicolumn{2}{l}{Overall ML Annot.}& .47&***& .37&***\\
		\bottomrule
	\end{tabular}
	\caption{Spearman correlation between human and automatic emotion annotation on events and chains. Significance levels are indicated as *: p < 0.05;  **: p < 0.01; ***: p < 0.001. ML Annot.: multi-label annotation.}
	\label{tab:corrclf}
\end{table}

\section{Predicted Emotion Analysis}\label{app:pred}
We compare the emotion label which is predicted with the highest likelihood score by our automatic emotion analysis method $e(c)$ to the emotion label each event chain has been created with $e(b{\mkern1.5mu{\cdot}\mkern1.5mu}s_5)$.
We evaluate this for each of our three data generation methods.
\Cref{tab:pe_tope_m_bc} shows the results for the Baseline approach.
\Cref{tab:pe_tope_m_gc} shows the results for our second approach, PC.
\Cref{tab:pe_tope_m_igc} shows the results for our third approach, PCR.

\begin{table*}
	\centering\small \setlength{\tabcolsep}{3pt}
	\begin{tabular}{l>{\raggedleft\arraybackslash}p{6.25mm}>{\raggedleft\arraybackslash}p{6.25mm}>{\raggedleft\arraybackslash}p{6.25mm}>{\raggedleft\arraybackslash}p{6.25mm}>{\raggedleft\arraybackslash}p{6.25mm}>{\raggedleft\arraybackslash}p{6.25mm}>{\raggedleft\arraybackslash}p{6.25mm}>{\raggedleft\arraybackslash}p{6.25mm}>{\raggedleft\arraybackslash}p{6.25mm}>{\raggedleft\arraybackslash}p{6.25mm}>{\raggedleft\arraybackslash}p{6.25mm}>{\raggedleft\arraybackslash}p{6.25mm}>{\raggedleft\arraybackslash}p{6.25mm} r}
		\toprule
		\diagbox[innerleftsep=0cm,width=1.25cm,height=1.4cm]{$e(b{\mkern1.5mu{\cdot}\mkern1.5mu}s_5)$}{$e(c)$}& \rotatebox[origin=c]{90}{Anger}&\rotatebox[origin=c]{90}{Boredom}&\rotatebox[origin=c]{90}{Disgust}&\rotatebox[origin=c]{90}{Fear}&\rotatebox[origin=c]{90}{Guilt}&\rotatebox[origin=c]{90}{Joy}&\rotatebox[origin=c]{90}{Pride}&\rotatebox[origin=c]{90}{Relief}&\rotatebox[origin=c]{90}{Sadness}&\rotatebox[origin=c]{90}{Shame}&\rotatebox[origin=c]{90}{Surprise}&\rotatebox[origin=c]{90}{Trust}&\rotatebox[origin=c]{90}{No-Emot.}&$\Sigma$\\
		\cmidrule(r){1-1}\cmidrule(rl){2-2}\cmidrule(rl){3-3}\cmidrule(rl){4-4}\cmidrule(rl){5-5}\cmidrule(rl){6-6}\cmidrule(rl){7-7}\cmidrule(rl){8-8}\cmidrule(rl){9-9}\cmidrule(rl){10-10}\cmidrule(rl){11-11}\cmidrule(rl){12-12}\cmidrule(rl){13-13}\cmidrule(rl){14-14}\cmidrule(l){15-15}
		Anger& \cellcolor{blue!27}278& \cellcolor{blue!1}11& \cellcolor{blue!2}28& \cellcolor{blue!4}44& \cellcolor{blue!1}12& \cellcolor{blue!9}95& \cellcolor{blue!3}34& \cellcolor{blue!20}201& \cellcolor{blue!11}118& \cellcolor{blue!3}33& \cellcolor{blue!12}124& \cellcolor{blue!1}11& \cellcolor{blue!1}11&1000\\
		Boredom& \cellcolor{blue!6}68& \cellcolor{blue!7}76& \cellcolor{blue!1}18& \cellcolor{blue!0}8& \cellcolor{blue!0}9& \cellcolor{blue!17}172& \cellcolor{blue!4}48& \cellcolor{blue!30}306& \cellcolor{blue!9}97& \cellcolor{blue!2}20& \cellcolor{blue!12}126& \cellcolor{blue!1}12& \cellcolor{blue!4}40&1000\\
		Disgust& \cellcolor{blue!22}221& \cellcolor{blue!1}16& \cellcolor{blue!16}167& \cellcolor{blue!4}42& \cellcolor{blue!2}26& \cellcolor{blue!6}60& \cellcolor{blue!2}21& \cellcolor{blue!17}173& \cellcolor{blue!7}78& \cellcolor{blue!3}39& \cellcolor{blue!11}116& \cellcolor{blue!1}14& \cellcolor{blue!2}27&1000\\
		Fear& \cellcolor{blue!5}55& \cellcolor{blue!0}2& \cellcolor{blue!0}5& \cellcolor{blue!28}288& \cellcolor{blue!1}18& \cellcolor{blue!8}81& \cellcolor{blue!2}28& \cellcolor{blue!26}264& \cellcolor{blue!6}64& \cellcolor{blue!4}42& \cellcolor{blue!12}120& \cellcolor{blue!1}18& \cellcolor{blue!1}15&1000\\
		Guilt& \cellcolor{blue!6}63& \cellcolor{blue!0}1& \cellcolor{blue!1}13& \cellcolor{blue!3}38& \cellcolor{blue!25}253& \cellcolor{blue!8}88& \cellcolor{blue!5}50& \cellcolor{blue!23}232& \cellcolor{blue!7}70& \cellcolor{blue!10}102& \cellcolor{blue!7}78& \cellcolor{blue!0}5& \cellcolor{blue!0}7&1000\\
		Joy& \cellcolor{blue!3}30& \cellcolor{blue!1}11& \cellcolor{blue!0}0& \cellcolor{blue!2}20& \cellcolor{blue!0}6& \cellcolor{blue!24}247& \cellcolor{blue!4}48& \cellcolor{blue!45}454& \cellcolor{blue!5}52& \cellcolor{blue!0}7& \cellcolor{blue!9}94& \cellcolor{blue!1}11& \cellcolor{blue!2}20&1000\\
		Pride& \cellcolor{blue!1}19& \cellcolor{blue!0}4& \cellcolor{blue!0}3& \cellcolor{blue!1}16& \cellcolor{blue!0}6& \cellcolor{blue!23}239& \cellcolor{blue!24}240& \cellcolor{blue!30}308& \cellcolor{blue!3}33& \cellcolor{blue!2}24& \cellcolor{blue!8}81& \cellcolor{blue!1}13& \cellcolor{blue!1}14&1000\\
		Relief& \cellcolor{blue!4}42& \cellcolor{blue!1}15& \cellcolor{blue!1}10& \cellcolor{blue!4}43& \cellcolor{blue!1}16& \cellcolor{blue!9}97& \cellcolor{blue!2}28& \cellcolor{blue!57}576& \cellcolor{blue!4}46& \cellcolor{blue!3}39& \cellcolor{blue!8}82& \cellcolor{blue!0}3& \cellcolor{blue!0}3&1000\\
		Sadness& \cellcolor{blue!7}72& \cellcolor{blue!1}10& \cellcolor{blue!1}13& \cellcolor{blue!3}36& \cellcolor{blue!1}19& \cellcolor{blue!12}126& \cellcolor{blue!4}49& \cellcolor{blue!17}173& \cellcolor{blue!32}320& \cellcolor{blue!2}24& \cellcolor{blue!11}115& \cellcolor{blue!1}19& \cellcolor{blue!2}24&1000\\
		Shame& \cellcolor{blue!7}79& \cellcolor{blue!0}3& \cellcolor{blue!0}9& \cellcolor{blue!3}34& \cellcolor{blue!11}110& \cellcolor{blue!6}68& \cellcolor{blue!4}46& \cellcolor{blue!18}189& \cellcolor{blue!10}105& \cellcolor{blue!23}230& \cellcolor{blue!10}105& \cellcolor{blue!0}6& \cellcolor{blue!1}16&1000\\
		Surprise& \cellcolor{blue!5}54& \cellcolor{blue!0}8& \cellcolor{blue!0}6& \cellcolor{blue!1}19& \cellcolor{blue!1}10& \cellcolor{blue!9}95& \cellcolor{blue!1}18& \cellcolor{blue!31}312& \cellcolor{blue!4}45& \cellcolor{blue!2}22& \cellcolor{blue!38}387& \cellcolor{blue!0}6& \cellcolor{blue!1}18&1000\\
		Trust& \cellcolor{blue!2}23& \cellcolor{blue!0}6& \cellcolor{blue!0}8& \cellcolor{blue!3}32& \cellcolor{blue!1}12& \cellcolor{blue!18}185& \cellcolor{blue!4}42& \cellcolor{blue!48}482& \cellcolor{blue!3}30& \cellcolor{blue!0}8& \cellcolor{blue!9}99& \cellcolor{blue!5}55& \cellcolor{blue!1}18&1000\\
		No-Emotion& \cellcolor{blue!4}48& \cellcolor{blue!3}39& \cellcolor{blue!1}18& \cellcolor{blue!2}24& \cellcolor{blue!0}8& \cellcolor{blue!12}123& \cellcolor{blue!3}37& \cellcolor{blue!40}401& \cellcolor{blue!9}91& \cellcolor{blue!2}22& \cellcolor{blue!12}123& \cellcolor{blue!1}16& \cellcolor{blue!5}50&1000\\
		$\Sigma$&1052&202&298&644&505&1676&689&4071&1149&612&1650&189&263&13000\\
		\bottomrule
	\end{tabular}
	\caption{Prompted emotion vs.\ top predicted emotion for event chains (generation method: Baseline).}
	\label{tab:pe_tope_m_bc}
\end{table*}

\begin{table*}
	\centering\small \setlength{\tabcolsep}{3pt}
	\begin{tabular}{l>{\raggedleft\arraybackslash}p{6.25mm}>{\raggedleft\arraybackslash}p{6.25mm}>{\raggedleft\arraybackslash}p{6.25mm}>{\raggedleft\arraybackslash}p{6.25mm}>{\raggedleft\arraybackslash}p{6.25mm}>{\raggedleft\arraybackslash}p{6.25mm}>{\raggedleft\arraybackslash}p{6.25mm}>{\raggedleft\arraybackslash}p{6.25mm}>{\raggedleft\arraybackslash}p{6.25mm}>{\raggedleft\arraybackslash}p{6.25mm}>{\raggedleft\arraybackslash}p{6.25mm}>{\raggedleft\arraybackslash}p{6.25mm}>{\raggedleft\arraybackslash}p{6.25mm} r}
		\toprule
		\diagbox[innerleftsep=0cm,width=1.25cm,height=1.4cm]{$e(b{\mkern1.5mu{\cdot}\mkern1.5mu}s_5)$}{$e(c)$}& \rotatebox[origin=c]{90}{Anger}&\rotatebox[origin=c]{90}{Boredom}&\rotatebox[origin=c]{90}{Disgust}&\rotatebox[origin=c]{90}{Fear}&\rotatebox[origin=c]{90}{Guilt}&\rotatebox[origin=c]{90}{Joy}&\rotatebox[origin=c]{90}{Pride}&\rotatebox[origin=c]{90}{Relief}&\rotatebox[origin=c]{90}{Sadness}&\rotatebox[origin=c]{90}{Shame}&\rotatebox[origin=c]{90}{Surprise}&\rotatebox[origin=c]{90}{Trust}&\rotatebox[origin=c]{90}{No-Emot.}&$\Sigma$\\
		\cmidrule(r){1-1}\cmidrule(rl){2-2}\cmidrule(rl){3-3}\cmidrule(rl){4-4}\cmidrule(rl){5-5}\cmidrule(rl){6-6}\cmidrule(rl){7-7}\cmidrule(rl){8-8}\cmidrule(rl){9-9}\cmidrule(rl){10-10}\cmidrule(rl){11-11}\cmidrule(rl){12-12}\cmidrule(rl){13-13}\cmidrule(rl){14-14}\cmidrule(l){15-15}
		Anger& \cellcolor{blue!18}182& \cellcolor{blue!2}20& \cellcolor{blue!2}20& \cellcolor{blue!3}31& \cellcolor{blue!0}6& \cellcolor{blue!11}116& \cellcolor{blue!3}33& \cellcolor{blue!19}198& \cellcolor{blue!14}147& \cellcolor{blue!2}22& \cellcolor{blue!17}177& \cellcolor{blue!1}19& \cellcolor{blue!2}29&1000\\
		Boredom& \cellcolor{blue!4}44& \cellcolor{blue!6}66& \cellcolor{blue!2}24& \cellcolor{blue!1}12& \cellcolor{blue!0}9& \cellcolor{blue!19}195& \cellcolor{blue!5}51& \cellcolor{blue!31}313& \cellcolor{blue!8}81& \cellcolor{blue!1}16& \cellcolor{blue!11}119& \cellcolor{blue!2}21& \cellcolor{blue!4}49&1000\\
		Disgust& \cellcolor{blue!14}148& \cellcolor{blue!2}23& \cellcolor{blue!17}176& \cellcolor{blue!3}34& \cellcolor{blue!3}32& \cellcolor{blue!10}105& \cellcolor{blue!2}24& \cellcolor{blue!15}157& \cellcolor{blue!6}65& \cellcolor{blue!2}24& \cellcolor{blue!14}142& \cellcolor{blue!2}26& \cellcolor{blue!4}44&1000\\
		Fear& \cellcolor{blue!4}44& \cellcolor{blue!1}16& \cellcolor{blue!1}10& \cellcolor{blue!15}151& \cellcolor{blue!2}22& \cellcolor{blue!15}156& \cellcolor{blue!2}28& \cellcolor{blue!25}259& \cellcolor{blue!6}63& \cellcolor{blue!2}29& \cellcolor{blue!15}158& \cellcolor{blue!2}20& \cellcolor{blue!4}44&1000\\
		Guilt& \cellcolor{blue!2}21& \cellcolor{blue!0}4& \cellcolor{blue!0}6& \cellcolor{blue!2}24& \cellcolor{blue!16}163& \cellcolor{blue!16}169& \cellcolor{blue!5}57& \cellcolor{blue!28}284& \cellcolor{blue!6}67& \cellcolor{blue!5}57& \cellcolor{blue!11}110& \cellcolor{blue!1}12& \cellcolor{blue!2}26&1000\\
		Joy& \cellcolor{blue!1}18& \cellcolor{blue!0}7& \cellcolor{blue!0}3& \cellcolor{blue!1}11& \cellcolor{blue!0}6& \cellcolor{blue!26}269& \cellcolor{blue!7}71& \cellcolor{blue!42}429& \cellcolor{blue!3}36& \cellcolor{blue!1}16& \cellcolor{blue!8}84& \cellcolor{blue!2}24& \cellcolor{blue!2}26&1000\\
		Pride& \cellcolor{blue!1}16& \cellcolor{blue!0}0& \cellcolor{blue!0}3& \cellcolor{blue!1}15& \cellcolor{blue!0}3& \cellcolor{blue!28}284& \cellcolor{blue!16}161& \cellcolor{blue!36}360& \cellcolor{blue!2}27& \cellcolor{blue!0}6& \cellcolor{blue!8}83& \cellcolor{blue!1}16& \cellcolor{blue!2}26&1000\\
		Relief& \cellcolor{blue!4}43& \cellcolor{blue!1}10& \cellcolor{blue!0}5& \cellcolor{blue!2}29& \cellcolor{blue!1}14& \cellcolor{blue!15}156& \cellcolor{blue!2}29& \cellcolor{blue!55}552& \cellcolor{blue!2}25& \cellcolor{blue!3}37& \cellcolor{blue!7}77& \cellcolor{blue!0}3& \cellcolor{blue!2}20&1000\\
		Sadness& \cellcolor{blue!3}34& \cellcolor{blue!2}22& \cellcolor{blue!1}11& \cellcolor{blue!1}14& \cellcolor{blue!1}10& \cellcolor{blue!21}210& \cellcolor{blue!6}62& \cellcolor{blue!17}174& \cellcolor{blue!23}230& \cellcolor{blue!1}15& \cellcolor{blue!14}149& \cellcolor{blue!1}16& \cellcolor{blue!5}53&1000\\
		Shame& \cellcolor{blue!2}25& \cellcolor{blue!0}3& \cellcolor{blue!0}3& \cellcolor{blue!2}27& \cellcolor{blue!8}83& \cellcolor{blue!11}112& \cellcolor{blue!6}62& \cellcolor{blue!28}285& \cellcolor{blue!8}86& \cellcolor{blue!16}160& \cellcolor{blue!11}111& \cellcolor{blue!1}12& \cellcolor{blue!3}31&1000\\
		Surprise& \cellcolor{blue!3}37& \cellcolor{blue!3}30& \cellcolor{blue!1}11& \cellcolor{blue!2}27& \cellcolor{blue!0}8& \cellcolor{blue!15}158& \cellcolor{blue!2}20& \cellcolor{blue!36}361& \cellcolor{blue!5}51& \cellcolor{blue!0}9& \cellcolor{blue!24}246& \cellcolor{blue!2}20& \cellcolor{blue!2}22&1000\\
		Trust& \cellcolor{blue!2}24& \cellcolor{blue!1}11& \cellcolor{blue!1}10& \cellcolor{blue!2}21& \cellcolor{blue!1}11& \cellcolor{blue!21}216& \cellcolor{blue!3}34& \cellcolor{blue!42}424& \cellcolor{blue!2}26& \cellcolor{blue!1}12& \cellcolor{blue!11}114& \cellcolor{blue!6}65& \cellcolor{blue!3}32&1000\\
		No-Emotion& \cellcolor{blue!2}21& \cellcolor{blue!2}28& \cellcolor{blue!0}7& \cellcolor{blue!1}14& \cellcolor{blue!0}8& \cellcolor{blue!21}218& \cellcolor{blue!5}51& \cellcolor{blue!41}415& \cellcolor{blue!5}52& \cellcolor{blue!0}8& \cellcolor{blue!9}97& \cellcolor{blue!1}19& \cellcolor{blue!6}62&1000\\
		$\Sigma$&657&240&289&410&375&2364&683&4211&956&411&1667&273&464&13000\\
		\bottomrule
	\end{tabular}
	\caption{Prompted emotion vs.\ top predicted emotion for event chains (generation method: PC).}
	\label{tab:pe_tope_m_gc}
\end{table*}

\begin{table*}
	\centering\small \setlength{\tabcolsep}{3pt}
	\begin{tabular}{l>{\raggedleft\arraybackslash}p{6.25mm}>{\raggedleft\arraybackslash}p{6.25mm}>{\raggedleft\arraybackslash}p{6.25mm}>{\raggedleft\arraybackslash}p{6.25mm}>{\raggedleft\arraybackslash}p{6.25mm}>{\raggedleft\arraybackslash}p{6.25mm}>{\raggedleft\arraybackslash}p{6.25mm}>{\raggedleft\arraybackslash}p{6.25mm}>{\raggedleft\arraybackslash}p{6.25mm}>{\raggedleft\arraybackslash}p{6.25mm}>{\raggedleft\arraybackslash}p{6.25mm}>{\raggedleft\arraybackslash}p{6.25mm}>{\raggedleft\arraybackslash}p{6.25mm} r}
		\toprule
		\diagbox[innerleftsep=0cm,width=1.25cm,height=1.4cm]{$e(b{\mkern1.5mu{\cdot}\mkern1.5mu}s_5)$}{$e(c)$}& \rotatebox[origin=c]{90}{Anger}&\rotatebox[origin=c]{90}{Boredom}&\rotatebox[origin=c]{90}{Disgust}&\rotatebox[origin=c]{90}{Fear}&\rotatebox[origin=c]{90}{Guilt}&\rotatebox[origin=c]{90}{Joy}&\rotatebox[origin=c]{90}{Pride}&\rotatebox[origin=c]{90}{Relief}&\rotatebox[origin=c]{90}{Sadness}&\rotatebox[origin=c]{90}{Shame}&\rotatebox[origin=c]{90}{Surprise}&\rotatebox[origin=c]{90}{Trust}&\rotatebox[origin=c]{90}{No-Emot.}&$\Sigma$\\
		\cmidrule(r){1-1}\cmidrule(rl){2-2}\cmidrule(rl){3-3}\cmidrule(rl){4-4}\cmidrule(rl){5-5}\cmidrule(rl){6-6}\cmidrule(rl){7-7}\cmidrule(rl){8-8}\cmidrule(rl){9-9}\cmidrule(rl){10-10}\cmidrule(rl){11-11}\cmidrule(rl){12-12}\cmidrule(rl){13-13}\cmidrule(rl){14-14}\cmidrule(l){15-15}
		Anger& \cellcolor{blue!25}257& \cellcolor{blue!1}12& \cellcolor{blue!3}30& \cellcolor{blue!3}35& \cellcolor{blue!0}6& \cellcolor{blue!4}47& \cellcolor{blue!3}30& \cellcolor{blue!17}172& \cellcolor{blue!19}194& \cellcolor{blue!2}28& \cellcolor{blue!15}154& \cellcolor{blue!0}8& \cellcolor{blue!2}27&1000\\
		Boredom& \cellcolor{blue!6}63& \cellcolor{blue!12}121& \cellcolor{blue!3}31& \cellcolor{blue!1}15& \cellcolor{blue!0}5& \cellcolor{blue!12}127& \cellcolor{blue!4}41& \cellcolor{blue!25}251& \cellcolor{blue!15}153& \cellcolor{blue!1}15& \cellcolor{blue!10}105& \cellcolor{blue!1}10& \cellcolor{blue!6}63&1000\\
		Disgust& \cellcolor{blue!21}219& \cellcolor{blue!0}4& \cellcolor{blue!25}257& \cellcolor{blue!5}55& \cellcolor{blue!3}37& \cellcolor{blue!5}56& \cellcolor{blue!2}22& \cellcolor{blue!10}101& \cellcolor{blue!7}74& \cellcolor{blue!3}31& \cellcolor{blue!8}89& \cellcolor{blue!1}18& \cellcolor{blue!3}37&1000\\
		Fear& \cellcolor{blue!4}44& \cellcolor{blue!0}5& \cellcolor{blue!1}12& \cellcolor{blue!29}292& \cellcolor{blue!2}22& \cellcolor{blue!7}72& \cellcolor{blue!1}17& \cellcolor{blue!26}261& \cellcolor{blue!7}72& \cellcolor{blue!5}51& \cellcolor{blue!10}101& \cellcolor{blue!1}10& \cellcolor{blue!4}41&1000\\
		Guilt& \cellcolor{blue!2}20& \cellcolor{blue!0}3& \cellcolor{blue!0}6& \cellcolor{blue!4}46& \cellcolor{blue!26}260& \cellcolor{blue!7}74& \cellcolor{blue!1}19& \cellcolor{blue!27}270& \cellcolor{blue!8}84& \cellcolor{blue!10}105& \cellcolor{blue!7}79& \cellcolor{blue!0}7& \cellcolor{blue!2}27&1000\\
		Joy& \cellcolor{blue!1}18& \cellcolor{blue!0}3& \cellcolor{blue!0}1& \cellcolor{blue!1}15& \cellcolor{blue!0}4& \cellcolor{blue!28}281& \cellcolor{blue!9}90& \cellcolor{blue!42}422& \cellcolor{blue!2}27& \cellcolor{blue!1}19& \cellcolor{blue!7}75& \cellcolor{blue!2}25& \cellcolor{blue!2}20&1000\\
		Pride& \cellcolor{blue!1}10& \cellcolor{blue!0}0& \cellcolor{blue!0}0& \cellcolor{blue!1}11& \cellcolor{blue!0}2& \cellcolor{blue!29}291& \cellcolor{blue!24}246& \cellcolor{blue!29}292& \cellcolor{blue!2}25& \cellcolor{blue!0}5& \cellcolor{blue!8}81& \cellcolor{blue!1}14& \cellcolor{blue!2}23&1000\\
		Relief& \cellcolor{blue!3}34& \cellcolor{blue!0}4& \cellcolor{blue!0}8& \cellcolor{blue!4}40& \cellcolor{blue!1}11& \cellcolor{blue!10}109& \cellcolor{blue!1}18& \cellcolor{blue!63}630& \cellcolor{blue!3}36& \cellcolor{blue!4}43& \cellcolor{blue!5}59& \cellcolor{blue!0}0& \cellcolor{blue!0}8&1000\\
		Sadness& \cellcolor{blue!3}30& \cellcolor{blue!0}5& \cellcolor{blue!0}6& \cellcolor{blue!2}20& \cellcolor{blue!0}6& \cellcolor{blue!13}137& \cellcolor{blue!5}56& \cellcolor{blue!11}118& \cellcolor{blue!43}433& \cellcolor{blue!2}22& \cellcolor{blue!11}113& \cellcolor{blue!1}11& \cellcolor{blue!4}43&1000\\
		Shame& \cellcolor{blue!1}17& \cellcolor{blue!0}1& \cellcolor{blue!0}3& \cellcolor{blue!4}43& \cellcolor{blue!10}108& \cellcolor{blue!5}55& \cellcolor{blue!2}27& \cellcolor{blue!24}248& \cellcolor{blue!10}102& \cellcolor{blue!28}280& \cellcolor{blue!9}92& \cellcolor{blue!0}6& \cellcolor{blue!1}18&1000\\
		Surprise& \cellcolor{blue!2}29& \cellcolor{blue!1}19& \cellcolor{blue!1}13& \cellcolor{blue!2}21& \cellcolor{blue!0}4& \cellcolor{blue!12}123& \cellcolor{blue!1}12& \cellcolor{blue!34}349& \cellcolor{blue!6}61& \cellcolor{blue!0}9& \cellcolor{blue!31}313& \cellcolor{blue!2}24& \cellcolor{blue!2}23&1000\\
		Trust& \cellcolor{blue!2}22& \cellcolor{blue!0}0& \cellcolor{blue!0}4& \cellcolor{blue!2}22& \cellcolor{blue!0}7& \cellcolor{blue!20}206& \cellcolor{blue!5}51& \cellcolor{blue!42}420& \cellcolor{blue!2}29& \cellcolor{blue!0}9& \cellcolor{blue!9}91& \cellcolor{blue!11}113& \cellcolor{blue!2}26&1000\\
		No-Emotion& \cellcolor{blue!2}21& \cellcolor{blue!4}42& \cellcolor{blue!0}6& \cellcolor{blue!1}15& \cellcolor{blue!0}4& \cellcolor{blue!12}128& \cellcolor{blue!7}70& \cellcolor{blue!40}405& \cellcolor{blue!7}77& \cellcolor{blue!1}13& \cellcolor{blue!10}104& \cellcolor{blue!3}31& \cellcolor{blue!8}84&1000\\
		$\Sigma$&784&219&377&630&476&1706&699&3939&1367&630&1456&277&440&13000\\
		\bottomrule
	\end{tabular}
	\caption{Prompted emotion vs.\ top predicted emotion for event chains (generation method: PCR).}
	\label{tab:pe_tope_m_igc}
\end{table*}

\section{Emotion Trajectories in Event Chains}\label{app:emotraj}
\begin{figure*}[]
	\centering
	\newcommand{\plotsaxisstyle}{
	width=6cm,
	xtick={1,2,3,4,5},
	ytick={0,0.1,...,1.0},
	yticklabels={0, 0.1, 0.2, 0.3, 0.4, 0.5, 0.6, 0.7, 0.8, 0.9, 1},
	ymin=-0.1,
	ymax=1.1,
	xmin=0.75,
	xmax=5.25,
	grid=major,
	domain=1:5,
	legend pos=north west,
	legend cell align={left},
	tick align=outside,
	tick pos=left,
}

\begin{minipage}{\textwidth}
	\centering
	\begin{subfigure}[t]{0.32\textwidth}
		\begin{tikzpicture}
			\begin{axis}[\plotsaxisstyle, tick label style={font=\scriptsize},legend style={font=\scriptsize},ylabel={$p(e=\text{Anger})$}, ylabel style={at={(.275,1.05)}, rotate=270, font=\scriptsize}]
				\addplot[thick, angercolor, mark=x] coordinates { (1,  0.120) (2,  0.267) (3,  0.296) (4,  0.325) (5,  0.244) };
				\addlegendentry{Anger Chains}
				\addplot[fill=angercolor, fill opacity=0.2, draw=none, forget plot] coordinates { (1,  0.421) (2,  0.676) (3,  0.715) (4,  0.753) (5,  0.617) (5,  0.000) (4,  0.000) (3,  0.000) (2,  0.000) (1,  0.000) };
				\addlegendentry{Events}
				\addplot[black, dashed] coordinates {(0.75,  0.015) (5.25,  0.015)};
			\end{axis}
		\end{tikzpicture}
	\end{subfigure}
	\begin{subfigure}[t]{0.32\textwidth}
		\begin{tikzpicture}
			\begin{axis}[\plotsaxisstyle, tick label style={font=\scriptsize},legend style={font=\scriptsize},ylabel={$p(e=\text{Boredom})$}, ylabel style={at={(.35,1.05)}, rotate=270, font=\scriptsize}]
				\addplot[thick, boredomcolor, mark=x] coordinates { (1,  0.093) (2,  0.094) (3,  0.110) (4,  0.137) (5,  0.119) };
				\addlegendentry{Boredom Chains}
				\addplot[fill=boredomcolor, fill opacity=0.2, draw=none, forget plot] coordinates { (1,  0.360) (2,  0.357) (3,  0.389) (4,  0.448) (5,  0.390) (5,  0.000) (4,  0.000) (3,  0.000) (2,  0.000) (1,  0.000) };
				\addlegendentry{Events}
				\addplot[black, dashed] coordinates {(0.75,  0.006) (5.25,  0.006)};
			\end{axis}
		\end{tikzpicture}
	\end{subfigure}
	\begin{subfigure}[t]{0.32\textwidth}
		\begin{tikzpicture}
			\begin{axis}[\plotsaxisstyle, tick label style={font=\scriptsize},legend style={font=\scriptsize},ylabel={$p(e=\text{Disgust})$}, ylabel style={at={(.31,1.05)}, rotate=270, font=\scriptsize}]
				\addplot[thick, disgustcolor, mark=x] coordinates { (1,  0.179) (2,  0.271) (3,  0.304) (4,  0.304) (5,  0.244) };
				\addlegendentry{Disgust Chains}
				\addplot[fill=disgustcolor, fill opacity=0.2, draw=none, forget plot] coordinates { (1,  0.538) (2,  0.679) (3,  0.719) (4,  0.716) (5,  0.605) (5,  0.000) (4,  0.000) (3,  0.000) (2,  0.000) (1,  0.000) };
				\addlegendentry{Events}
				\addplot[black, dashed] coordinates {(0.75,  0.002) (5.25,  0.002)};
			\end{axis}
		\end{tikzpicture}
	\end{subfigure}
\end{minipage}
\begin{minipage}{\textwidth}
	\centering
	\begin{subfigure}[t]{0.32\textwidth}
		\begin{tikzpicture}
			\begin{axis}[\plotsaxisstyle, tick label style={font=\scriptsize},legend style={font=\scriptsize},ylabel={$p(e=\text{Fear})$}, ylabel style={at={(.225,1.05)}, rotate=270, font=\scriptsize}]
				\addplot[thick, fearcolor, mark=x] coordinates { (1,  0.355) (2,  0.403) (3,  0.429) (4,  0.532) (5,  0.281) };
				\addlegendentry{Fear Chains}
				\addplot[fill=fearcolor, fill opacity=0.2, draw=none, forget plot] coordinates { (1,  0.800) (2,  0.858) (3,  0.888) (4,  0.993) (5,  0.684) (5,  0.000) (4,  0.070) (3,  0.000) (2,  0.000) (1,  0.000) };
				\addlegendentry{Events}
				\addplot[black, dashed] coordinates {(0.75,  0.008) (5.25,  0.008)};
			\end{axis}
		\end{tikzpicture}
	\end{subfigure}
	\begin{subfigure}[t]{0.32\textwidth}
		\begin{tikzpicture}
			\begin{axis}[\plotsaxisstyle, tick label style={font=\scriptsize},legend style={font=\scriptsize},ylabel={$p(e=\text{Guilt})$}, ylabel style={at={(.25,1.05)}, rotate=270, font=\scriptsize}]
				\addplot[thick, guiltcolor, mark=x] coordinates { (1,  0.127) (2,  0.257) (3,  0.313) (4,  0.340) (5,  0.242) };
				\addlegendentry{Guilt Chains}
				\addplot[fill=guiltcolor, fill opacity=0.2, draw=none, forget plot] coordinates { (1,  0.427) (2,  0.652) (3,  0.723) (4,  0.760) (5,  0.600) (5,  0.000) (4,  0.000) (3,  0.000) (2,  0.000) (1,  0.000) };
				\addlegendentry{Events}
				\addplot[black, dashed] coordinates {(0.75,  0.000) (5.25,  0.000)};
			\end{axis}
		\end{tikzpicture}
	\end{subfigure}
	\begin{subfigure}[t]{0.32\textwidth}
		\begin{tikzpicture}
			\begin{axis}[\plotsaxisstyle, tick label style={font=\scriptsize},legend style={font=\scriptsize},ylabel={$p(e=\text{Joy})$}, ylabel style={at={(.215,1.05)}, rotate=270, font=\scriptsize}]
				\addplot[thick, joycolor, mark=x] coordinates { (1,  0.081) (2,  0.089) (3,  0.122) (4,  0.154) (5,  0.272) };
				\addlegendentry{Joy Chains}
				\addplot[fill=joycolor, fill opacity=0.2, draw=none, forget plot] coordinates { (1,  0.328) (2,  0.345) (3,  0.412) (4,  0.467) (5,  0.662) (5,  0.000) (4,  0.000) (3,  0.000) (2,  0.000) (1,  0.000) };
				\addlegendentry{Events}
				\addplot[black, dashed] coordinates {(0.75,  0.327) (5.25,  0.327)};
			\end{axis}
		\end{tikzpicture}
	\end{subfigure}
\end{minipage}
\begin{minipage}{\textwidth}
	\centering
	\begin{subfigure}[t]{0.32\textwidth}
		\begin{tikzpicture}
			\begin{axis}[\plotsaxisstyle, tick label style={font=\scriptsize},legend style={font=\scriptsize},ylabel={$p(e=\text{Pride})$}, ylabel style={at={(.25,1.05)}, rotate=270, font=\scriptsize}]
				\addplot[thick, pridecolor, mark=x] coordinates { (1,  0.236) (2,  0.263) (3,  0.295) (4,  0.353) (5,  0.243) };
				\addlegendentry{Pride Chains}
				\addplot[fill=pridecolor, fill opacity=0.2, draw=none, forget plot] coordinates { (1,  0.615) (2,  0.645) (3,  0.688) (4,  0.761) (5,  0.578) (5,  0.000) (4,  0.000) (3,  0.000) (2,  0.000) (1,  0.000) };
				\addlegendentry{Events}
				\addplot[black, dashed] coordinates {(0.75,  0.014) (5.25,  0.014)};
			\end{axis}
		\end{tikzpicture}
	\end{subfigure}
	\begin{subfigure}[t]{0.32\textwidth}
		\begin{tikzpicture}
			\begin{axis}[\plotsaxisstyle, tick label style={font=\scriptsize},legend style={font=\scriptsize},ylabel={$p(e=\text{Relief})$}, ylabel style={at={(.275,1.05)}, rotate=270, font=\scriptsize}]
				\addplot[thick, reliefcolor, mark=x] coordinates { (1,  0.072) (2,  0.140) (3,  0.208) (4,  0.324) (5,  0.612) };
				\addlegendentry{Relief Chains}
				\addplot[fill=reliefcolor, fill opacity=0.2, draw=none, forget plot] coordinates { (1,  0.301) (2,  0.458) (3,  0.591) (4,  0.767) (5,  1.000) (5,  0.178) (4,  0.000) (3,  0.000) (2,  0.000) (1,  0.000) };
				\addlegendentry{Events}
				\addplot[black, dashed] coordinates {(0.75,  0.083) (5.25,  0.083)};
			\end{axis}
		\end{tikzpicture}
	\end{subfigure}
	\begin{subfigure}[t]{0.32\textwidth}
		\begin{tikzpicture}
			\begin{axis}[\plotsaxisstyle, tick label style={font=\scriptsize},legend style={font=\scriptsize},ylabel={$p(e=\text{Sadness})$}, ylabel style={at={(.32,1.05)}, rotate=270, font=\scriptsize}]
				\addplot[thick, sadnesscolor, mark=x] coordinates { (1,  0.385) (2,  0.455) (3,  0.497) (4,  0.499) (5,  0.418) };
				\addlegendentry{Sadness Chains}
				\addplot[fill=sadnesscolor, fill opacity=0.2, draw=none, forget plot] coordinates { (1,  0.847) (2,  0.921) (3,  0.961) (4,  0.959) (5,  0.856) (5,  0.000) (4,  0.038) (3,  0.033) (2,  0.000) (1,  0.000) };
				\addlegendentry{Events}
				\addplot[black, dashed] coordinates {(0.75,  0.013) (5.25,  0.013)};
			\end{axis}
		\end{tikzpicture}
	\end{subfigure}
\end{minipage}
\begin{minipage}{\textwidth}
	\centering
	\begin{subfigure}[t]{0.32\textwidth}
		\begin{tikzpicture}
			\begin{axis}[\plotsaxisstyle, tick label style={font=\scriptsize},legend style={font=\scriptsize},ylabel={$p(e=\text{Shame})$}, ylabel style={at={(.29,1.05)}, rotate=270, font=\scriptsize}]
				\addplot[thick, shamecolor, mark=x] coordinates { (1,  0.091) (2,  0.217) (3,  0.269) (4,  0.321) (5,  0.268) };
				\addlegendentry{Shame Chains}
				\addplot[fill=shamecolor, fill opacity=0.2, draw=none, forget plot] coordinates { (1,  0.333) (2,  0.574) (3,  0.655) (4,  0.731) (5,  0.640) (5,  0.000) (4,  0.000) (3,  0.000) (2,  0.000) (1,  0.000) };
				\addlegendentry{Events}
				\addplot[black, dashed] coordinates {(0.75,  0.002) (5.25,  0.002)};
			\end{axis}
		\end{tikzpicture}
	\end{subfigure}
	\begin{subfigure}[t]{0.32\textwidth}
		\begin{tikzpicture}
			\begin{axis}[\plotsaxisstyle, tick label style={font=\scriptsize},legend style={font=\scriptsize},ylabel={$p(e=\text{Surprise})$}, ylabel style={at={(.325,1.05)}, rotate=270, font=\scriptsize}]
				\addplot[thick, surprisecolor, mark=x] coordinates { (1,  0.164) (2,  0.116) (3,  0.119) (4,  0.152) (5,  0.304) };
				\addlegendentry{Surprise Chains}
				\addplot[fill=surprisecolor, fill opacity=0.2, draw=none, forget plot] coordinates { (1,  0.470) (2,  0.370) (3,  0.386) (4,  0.458) (5,  0.695) (5,  0.000) (4,  0.000) (3,  0.000) (2,  0.000) (1,  0.000) };
				\addlegendentry{Events}
				\addplot[black, dashed] coordinates {(0.75,  0.138) (5.25,  0.138)};
			\end{axis}
		\end{tikzpicture}
	\end{subfigure}
	\begin{subfigure}[t]{0.32\textwidth}
		\begin{tikzpicture}
			\begin{axis}[\plotsaxisstyle, tick label style={font=\scriptsize},legend style={font=\scriptsize},ylabel={$p(e=\text{Trust})$}, ylabel style={at={(.26,1.05)}, rotate=270, font=\scriptsize}]
				\addplot[thick, trustcolor, mark=x] coordinates { (1,  0.069) (2,  0.127) (3,  0.142) (4,  0.194) (5,  0.130) };
				\addlegendentry{Trust Chains}
				\addplot[fill=trustcolor, fill opacity=0.2, draw=none, forget plot] coordinates { (1,  0.280) (2,  0.401) (3,  0.434) (4,  0.520) (5,  0.395) (5,  0.000) (4,  0.000) (3,  0.000) (2,  0.000) (1,  0.000) };
				\addlegendentry{Events}
				\addplot[black, dashed] coordinates {(0.75,  0.025) (5.25,  0.025)};
			\end{axis}
		\end{tikzpicture}
	\end{subfigure}
\end{minipage}
	\caption{Mean and standard deviation of emotion trajectories in event chains (x-axes corresponding to the first n sentences of the chains; generation method: PCR) in comparison to mean emotion of events in isolation (dashed lines).}
	\label{fig:allemotraj}
\end{figure*}
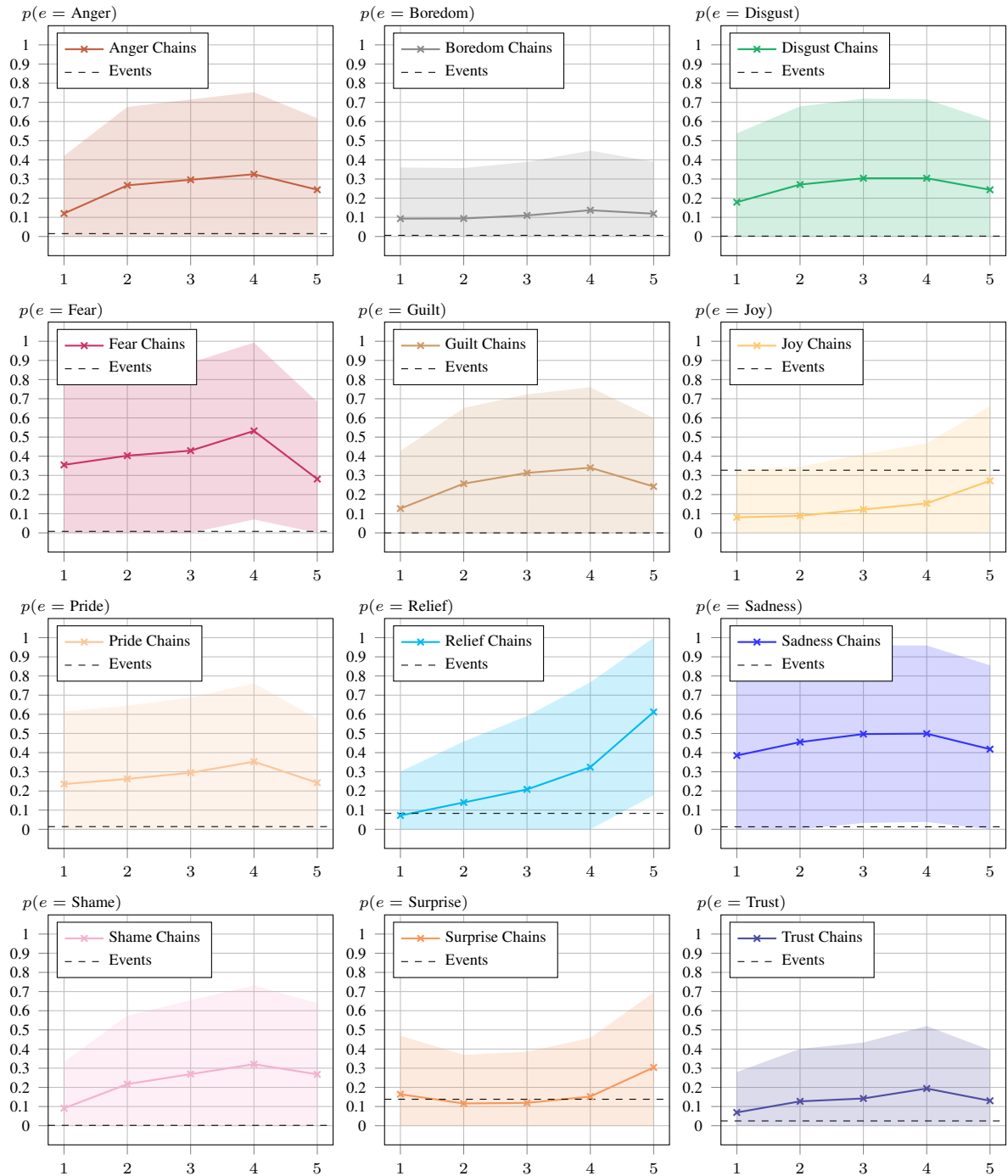

Based on our automatic emotion analysis, we evaluate the mean and standard deviation of emotion trajectories in event chains generated by PCR in comparison to mean emotion of events in isolation.
Plots how the trajectories of the different emotions develop over the sentences of the narratives are shown in \Cref{fig:allemotraj}.
The colors used for the plots of the different emotions correspond to the ones used in \Cref{fig:emotraj}.

\section{Correlation of Coherence and Emotion Uncertainty}\label{app:cohcorr}

Our analysis shows that the possibility of disambiguating events in emotion analysis depends on the particular emotion we want to trigger through contextual narratives.
Now, we want to find out if emotion properties of the initial events already indicate whether disambiguation is possible through coherent narratives. 
Therefore, we investigate the relationship between model uncertainty in emotion analysis and the coherence of the generated backstories.
We quantify uncertainty using the entropy of the predicted probability distribution.

To examine a potential correlation with the average coherence score, we calculate the Pearson ($r^{(p)}$) and Spearman ($r^{(s)}$) correlation of emotion model uncertainty $e$, i.e.\ the entropy of the probability distribution of zero-shot emotion analysis on the events, with average coherence $c$ of the corresponding event chains.
The correlation scores of events to the event chains generated through the various methods are as follows.
\begin{compactitem}
	\item Baseline: $r^{(p)} =-.09$  (**),  $r^{(s)}=-.10$ (**)
	\item PC: $r^{(p)} =-.08$  (*),  $r^{(s)}=-.11$ (***)
	\item PCR: $r^{(p)} =-.13$  (***),  $r^{(s)}=-.16$ (***)
\end{compactitem}

The results reveal a significant but modest negative correlation between the uncertainty of emotion predictions for the events and the coherence of the corresponding event chains.
This finding implies that when the model exhibits uncertainty regarding the emotion impact of an event, such uncertainty may extend to the generation of narratives.
Specifically, high entropy indicates that the event may lend itself to multiple interpretations or emotion responses.
Consequently, narratives associated with such events may allow for variability in their placement within the storyline, leading to lower coherence scores assessed through the shuffle test.
We therefore find that if the model shows low uncertainty in emotion analysis for events, this indicates that these events are easier to disambiguate using coherent narratives.

\end{document}